\documentclass[letterpaper]{article} 
\usepackage{aaai2027}
\usepackage[hyphens]{url} 
\usepackage{graphicx} 
\def\UrlFont{\rm} 
\usepackage{natbib} 
\usepackage{caption} 
\usepackage{amsmath}
\usepackage{amssymb}
\usepackage{booktabs}
\usepackage{multirow}
\usepackage{tabularx}

\title{PolyBridgeBench: Benchmarking Multimodal LLMs for Physics-Grounded Bridge Design}
\author{
    Zicheng Zhao\textsuperscript{\rm 1},
    Dongyin Chen\textsuperscript{\rm 1},
    Rui Xu\textsuperscript{\rm 1,2},
    Yinghui Xu\textsuperscript{\rm 1}\corresponding
}
\affiliations{
    \textsuperscript{\rm 1}Fudan University\\
    \textsuperscript{\rm 2}Shanghai Innovation Institute\\
    25213050507@m.fudan.edu.cn, xuyinghui@fudan.edu.cn
}

\begin{document}

\maketitle

\begin{abstract}
Multimodal large language models, or MLLMs, perform well at visual understanding and structured generation, yet these capabilities do not establish whether an engineering design will work when executed. Existing benchmarks assess spatial reasoning, structural validity, or physics-grounded construction, but they do not determine whether MLLMs can synthesize complete load-bearing structures and repair them after simulator execution exposes a failure. We introduce PolyBridgeBench, an executable benchmark for multimodal bridge design. A model receives a visual scene and structured engineering constraints and generates a complete node--member--material topology. Deterministic legality checks gate execution in a native dynamic physics simulation. Following an execution failure, the benchmark returns temporal visual evidence from the failed rollout and evaluates repair under a fixed interaction budget. Separate measurements of deterministic validity, dynamic functional success, and post-failure recovery identify the stage at which design fails. Experiments with six representative MLLMs across 189 levels expose a substantial gap between deterministic validity and dynamic success, pronounced sensitivity to material budgets, and limited post-failure recovery under the primary strict-budget setting.
\end{abstract}


\begin{figure*}[t]
    \centering
    \includegraphics[width=\textwidth]{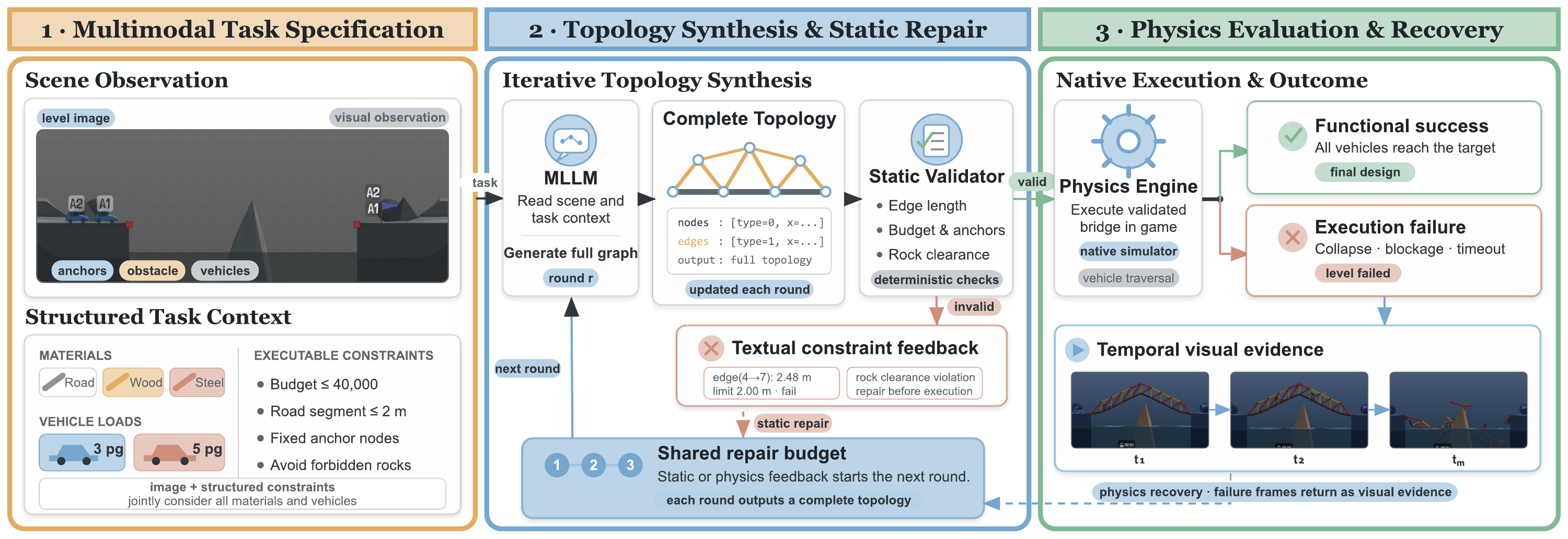}
    \caption{Overview of PolyBridgeBench. The input combines a visual scene with structured materials, loads, geometry, and budget constraints. Each model call emits a complete topology. Invalid candidates receive textual constraint feedback and return to synthesis without entering physics; valid candidates are executed in the native simulator. A physical failure returns temporal frames from that rollout, whereas functional success requires all vehicles to reach their targets. Both repair paths share a fixed interaction budget.}
    \label{fig:overview}
\end{figure*}

\section{Introduction}

A bridge may obey every explicit construction rule and still collapse when a vehicle enters it. An engineering artifact is useful only if it performs its intended function under execution, not merely if it conforms to a symbolic specification. The central evaluation question is whether an MLLM can translate a multimodal scene and engineering constraints into a complete structure that works under native dynamics, then revise that structure when execution exposes a flaw.

Existing benchmarks expose relevant capabilities but stop short of this joint evaluation. Spatial and physical reasoning suites probe compositional relations, metric scene understanding, event prediction, and action selection~\cite{clevr,spatialvlm,vsibench,phyre,physion,physbench}. ORIGAMISPACE evaluates multi-step spatial reasoning under explicit mathematical constraints~\cite{origamispace}, DreamHouse and BrDSL study structural compliance and analysis~\cite{dreamhouse,brdsl}, and BuildArena evaluates language-driven construction in a physics-grounded environment~\cite{buildarena}. No existing protocol follows a candidate from complete structure synthesis through separate legality and functionality tests to recovery based on evidence from its own failed execution.

Reference similarity cannot define correctness because structural design is one-to-many: a level may admit several functionally valid topologies. This limitation also motivates executable, test-based CAD evaluation~\cite{cadcodeverify,cadtestbench}. Legality must also remain distinct from functionality. Deterministic rules can verify connectivity, geometry, materials, and cost, but they cannot establish whether a legal topology will survive dynamic loading and permit traversal. Repair introduces a further measurement problem. Without fixed limits on model calls and physics attempts, higher final success may merely reflect more test-time computation. Diagnostic embodied-agent benchmarks likewise show that final success alone cannot localize the missing capability~\cite{embodiedagentinterface}. Bridge construction concentrates these evaluation problems in a controlled setting with dense constraints, non-unique solutions, and failures that unfold over time.

We therefore evaluate a generated structure as an executable hypothesis rather than a static answer. The benchmark accepts any design that is legal and works under native dynamics without matching it to a reference bridge. When execution fails, evidence from that same rollout supports a revision within the remaining budget. This staged protocol admits multiple valid solutions, measures the gap between compliance and functionality directly, and makes repair cost explicit.

PolyBridgeBench instantiates this evaluation across 189 levels from 27 structural families. A model generates a complete node--member--material topology from a scene image and structured engineering constraints. Each candidate undergoes deterministic validation before serialization into Poly Bridge for native physics execution. After a failure, the model may revise the full topology under fixed budgets for model calls and physics attempts; the post-failure input includes temporal visual evidence from the rollout. The protocol records legality, physics success, and recovery separately. We evaluate six representative MLLMs under strict and loose material budgets and analyze their performance across parameterized and compositional structural conditions.

This work makes the following contributions:
\begin{itemize}
    \item We introduce PolyBridgeBench, an executable benchmark that formulates multimodal bridge construction as complete topology generation under explicit engineering constraints and native dynamic execution.
    \item We develop a layered diagnostic protocol that measures deterministic validity, single-call and budgeted dynamic success, and post-failure recovery separately. The resulting measurements identify where rule-compliant designs cease to function.
    \item Evaluation across 189 levels and six MLLMs exposes a substantial gap between deterministic validity and dynamic success, pronounced sensitivity to material budgets, and distinct model-specific failure bottlenecks.
\end{itemize}

\begin{figure*}[t]
    \centering
    \includegraphics[width=\textwidth]{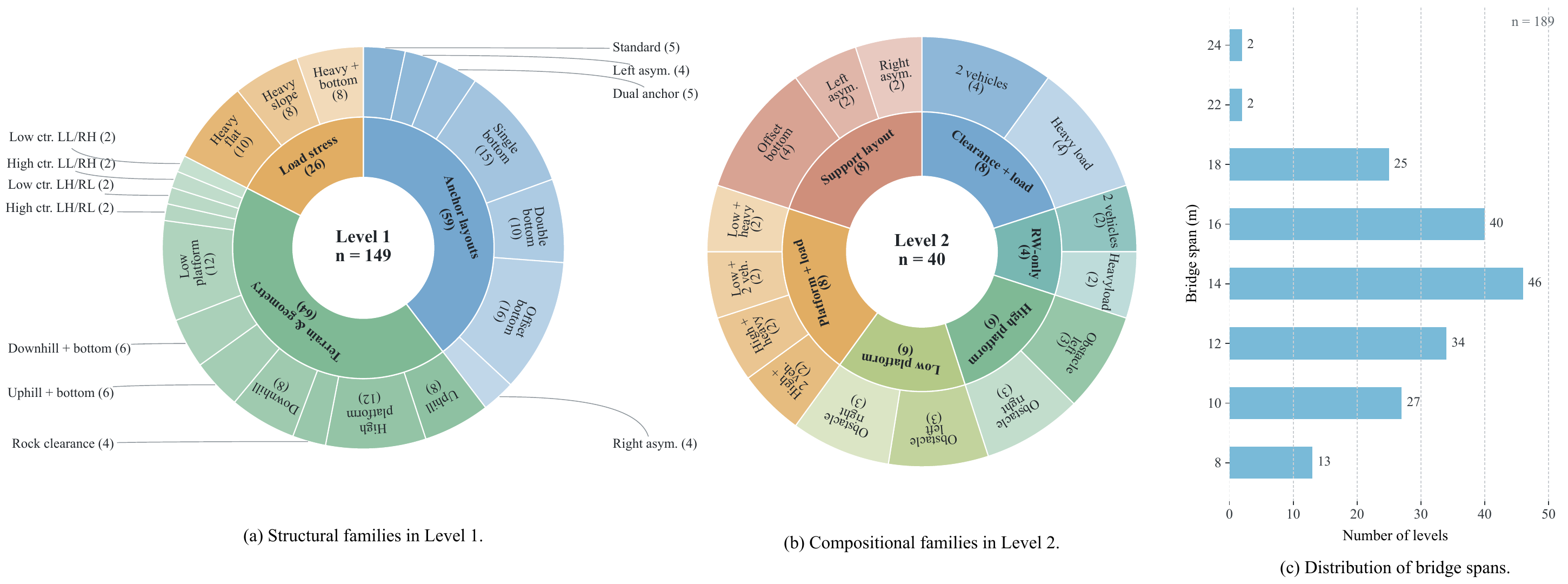}
    \caption{Composition of PolyBridgeBench. Panel a shows the 149 Core Structural Suite levels from 21 parameterized structural families. Panel b shows the 40 Compositional Challenge Suite levels from six families that combine multiple engineering constraints. Panel c reports the bridge-span distribution over all 189 levels. Sector areas and bar lengths are proportional to level counts. The suites are complementary evaluation strata, and all six models are evaluated on all 189 levels.}
    \label{fig:benchmark-composition}
\end{figure*}

\section{Related Work}

\subsection{Multimodal Spatial and Physical Reasoning}

CLEVR, NLVR2, and StepGame established controlled tests of compositional and spatial relations~\cite{clevr,nlvr2,stepgame}. Newer benchmarks cover metric estimation, reference-frame grounding, video-based spatial memory, and sequential assembly~\cite{spatialvlm,robospatial,vsibench,mindgap,legopuzzles}. ORIGAMISPACE adds foldability constraints and crease-pattern generation~\cite{origamispace}. These benchmarks expose failures hidden by static visual question answering, but their outputs remain answers, plans, predictions, or compilable geometry rather than load-bearing topologies whose functionality requires dynamic execution.

Physical-reasoning benchmarks test commonsense, object attributes, event prediction, or intervention in predefined scenes~\cite{piqa,newton,phyre,physion,physbench}; LLMPhy uses a simulator for parameter estimation and dynamics prediction~\cite{llmphy}. They assess existing scenes rather than artifact synthesis. PolyBridgeBench instead requires creating the artifact that determines subsequent dynamics, using execution as both the functional criterion and evidence for revision.

\subsection{Constrained Structural Generation and Engineering Analysis}

Executable design pipelines directly verify structured artifacts. CAD-Assistant executes and adapts VLLM-planned FreeCAD actions~\cite{cadassistant}; CADCodeVerify uses rendered execution feedback~\cite{cadcodeverify}; Text2BIM combines programmatic authoring with model checking~\cite{text2bim}; and CADTestBench tests geometric and topological requirements~\cite{cadtestbench}. DreamHouse checks timber-frame geometry, structure, constructability, and code compliance, while BrDSL connects bridge specifications to finite-element analysis~\cite{dreamhouse,brdsl}. These criteria emphasize compilation, compliance, or scoped analysis. PolyBridgeBench instead accepts any legal topology that completes vehicle traversal under native dynamics.

\subsection{Executable Engineering Construction and Interactive Evaluation}

Language-model agents interleave reasoning with actions, while PlanBench and Embodied Agent Interface expose planning or diagnostic failures beyond final success~\cite{react,planbench,embodiedagentinterface}. BuildArena is the closest physics-grounded construction benchmark, evaluating language-driven 3D assembly through transport, support, and lift tasks~\cite{buildarena}. PolyBridgeBench changes the evaluation unit from an action sequence to a complete graph: validation and native execution expose distinct failures, and evidence from a failed rollout tests whether a subsequent graph recovers within the remaining budget.

\section{PolyBridgeBench}

Figure~\ref{fig:overview} shows two diagnostic paths under one call budget. Invalid graphs, $D(G_t;x)=0$, receive validator feedback without consuming a physics attempt; valid but unsuccessful graphs, $D(G_t;x)=1$ and $P(G_t;x)=0$, receive rollout evidence. Both paths require a new complete topology.

\subsection{Task Formulation}

Each level is $x=(I_0,T_0,C,E)$, comprising an initial editor image, native-anchor topology, task constraints (materials, budget, vehicles, and loads), and environment constraints (terrain and clearance regions). At call $t$, the model receives $I_t$, $T_{t-1}$, fixed conditions $(C,E)$, and prior feedback $H_{<t}$, then outputs
\begin{equation}
G_t=(V_t,M_t),
\end{equation}
where $V_t$ contains nodes with coordinates and kinematic attributes, while $M_t$ contains members with endpoints, materials, and connectivity. Every call returns the complete structure rather than recommendations or incremental edits, enabling independent parsing, validation, execution, and replay.

Within a fixed interaction budget, the objective is a design satisfying legality $D(G_t;x)=1$ and dynamic functionality $P(G_t;x)=1$, where functionality requires vehicle traversal under native physics. This accepts multiple working topologies rather than one reference bridge.

\subsection{Benchmark Composition}

PolyBridgeBench contains 189 executable levels from 27 procedural families. The 149-level Core Structural Suite varies span, slope, bank geometry, anchors, platforms, materials, vehicles, and loads. The 40-level Compositional Challenge Suite combines clearance, material, load, vehicle, platform, and support constraints to test performance when several requirements apply at once.

Figure~\ref{fig:benchmark-composition} visualizes the benchmark composition across structural families, compositional subtypes, and bridge spans.

The suites are complementary strata, not an in-/out-of-distribution split; every model runs on all levels. Core measures fundamental variations, while Compositional isolates combined constraints.

The generator retains family identifiers, parameters, and tags for offline analysis but hides them from models. Inputs expose only native anchors, available materials, vehicle configuration, applicable clearance regions, and a per-level budget calibrated from a manually built reference design. The primary evaluation enforces this budget as a hard constraint.

\subsection{Structured Bridge Representation}

Bridges use node--member--material graphs. Nodes store identifiers, 2D coordinates, and fixed-anchor status; members store endpoints, materials, and save-format geometry. Native anchors are immutable, generated nodes are non-kinematic, and materials have distinct length limits and costs; each level may further restrict admissible materials. The graph supports diverse topologies, deterministic parsing, and direct game-save serialization; the appendix provides the schema, coordinate conventions, and prompt.

\subsection{Deterministic Validation and Dynamic Execution}

Before physics, deterministic validation checks complete topology, immutable anchors, valid references, anchor-connected dynamic components, member lengths, materials, cost, duplicate edges, isolated nodes, and forbidden-region crossings. It defines the legal action space but neither completes designs nor predicts stability.

Invalid candidates consume a call but no physics attempt and receive a deterministic textual error. Valid candidates are serialized and executed in the native simulator. A unified detector checks whether all vehicles complete their required traversal within a fixed window; otherwise the run is a physical failure or timeout without an inferred cause. Topologies, responses, frames, and resource records are retained for audit.


After physical failure, selected rollout frames, failure status, and the current topology form the next observation. The model regenerates a complete graph within the remaining three-call/three-execution budget; an episode may include validator repair, physical failures, and later success.

Frames provide observational evidence, not an automatic diagnosis: selection removes repeated post-termination frames but labels no causal breakpoint. Together with failure status and the current topology, they form the standard post-failure observation. Recovery therefore measures this complete feedback package rather than any single modality.

\section{Evaluation Protocol}

\subsection{Layered Metrics}

Let the benchmark contain $N$ levels. For level $i$, let $v_i^t\in\{0,1\}$ indicate whether the candidate from the $t$-th LLM call passes deterministic validation. If that candidate enters physics execution, $y_i^t\in\{0,1\}$ indicates whether it succeeds functionally. We define
\begin{equation}
\mathrm{Valid@}k=
\frac{1}{N}\sum_{i=1}^{N}
\mathbb{I}\!\left[\exists t\leq k,\ v_i^t=1\right],
\end{equation}
which measures whether the model produces at least one deterministically valid topology within $k$ calls. We measure end-to-end physical success as
\begin{equation}
\mathrm{Success@}k=
\frac{1}{N}\sum_{i=1}^{N}
\mathbb{I}\!\left[\exists t\leq k,\ y_i^t=1\right].
\end{equation}
Success@1 requires the first model output to pass validation and succeed on its first physics execution. If the model never produces a valid design, the level cannot enter physics execution and is counted as unsuccessful for Success@k.

To quantify recovery after execution failure, let $\mathcal{R}_k$ contain the levels that experience a physical failure before the $k$-th call and retain at least one LLM call and one physics attempt after that failure. We define
\begin{equation}
\mathrm{Recovery@}k=
\frac{1}{|\mathcal{R}_k|}
\sum_{i\in\mathcal{R}_k}
\mathbb{I}\!\left[
\exists t_f<t_s\leq k,\
y_i^{t_f}=0\land y_i^{t_s}=1
\right].
\end{equation}
Recovery@k is computed per level rather than per failure event. A trajectory of $F\!\rightarrow\!F\!\rightarrow\!S$ therefore counts as one successful recovery, with no additional weight for the second failed attempt. Because the conditional denominator can differ across models, Recovery@3 is interpreted as a conditional diagnostic rather than another aggregate success rate. Parsing failures, validator-error categories, physics attempts, and model calls provide additional evidence about where failures occur.

\subsection{Models and Unified Settings}

The primary evaluation covers six representative MLLMs: Gemini-3-Flash-Preview, Qwen3-VL-Plus, Gemma-4-31B-IT, Kimi-K2.5, GPT-5.2, and Claude-Sonnet-5. Every model receives semantically equivalent inputs under the same topology-output specification. The appendix documents each model's thinking or reasoning controls, exact API snapshot, access date, and maximum output length.

Each episode permits at most three LLM calls and three physics attempts. We set the temperature to 0.2 when the provider exposes this control; otherwise, we retain the provider default and report it in the appendix. Physics execution uses a 12-second simulation window and samples raw frames at 4 fps. Four frames from each failed rollout form a contact sheet for temporal visual input. Experiments use individually calibrated per-level budgets under strict enforcement and a looser uniform-budget setting.

\section{Experiments and Analysis}

The analysis is organized around three research questions. RQ1 asks whether current MLLMs can generate executable bridges and recover under bounded interaction. RQ2 locates failures in structured generation, deterministic legality, or dynamic functionality. RQ3 examines the material and interaction costs of successful designs.

\begin{table*}[t]
\centering
\begin{tabularx}{\textwidth}{@{}l >{\centering\arraybackslash}X >{\centering\arraybackslash}X >{\centering\arraybackslash}X >{\centering\arraybackslash}X >{\centering\arraybackslash}X >{\centering\arraybackslash}X >{\centering\arraybackslash}X >{\centering\arraybackslash}X >{\centering\arraybackslash}X @{}}
\toprule
\multirow{2}{*}{\textbf{Model}} & \multicolumn{2}{c}{\textbf{Level 1}} & \multicolumn{2}{c}{\textbf{Level 2}} & \multicolumn{5}{c}{\textbf{Overall}} \\
\cmidrule(lr){2-3}\cmidrule(lr){4-5}\cmidrule(lr){6-10}
& \textbf{S@1} & \textbf{S@3} & \textbf{S@1} & \textbf{S@3} & \textbf{V@1} & \textbf{S@1} & \textbf{V@3} & \textbf{S@3} & \textbf{R@3} \\
\midrule
\multicolumn{10}{c}{\textit{Loose budget}} \\
\midrule
gemini-3-flash-preview & \textbf{80.5} & \textbf{92.6} & \textbf{40.0} & \textbf{55.0} & 84.7 & \textbf{72.0} & 96.3 & \textbf{84.7} & 33.3 \\
qwen3-vl-plus          & 3.4  & 15.4 & 0.0  & 0.0  & 50.8 & 2.6  & 92.1 & 12.2 & 10.8 \\
gemma-4-31b-it         & 22.8 & 57.7 & 0.0  & 2.5  & 31.2 & 18.0 & 79.9 & 46.0 & 37.9 \\
kimi-k2.5              & 7.4  & 23.5 & 0.0  & 0.0  & 12.2 & 5.8  & 41.8 & 18.5 & 28.0 \\
gpt-5.2                & 7.4  & 43.0 & 0.0  & 0.0  & 10.6 & 5.8  & 66.1 & 33.9 & 25.9 \\
claude-sonnet-5        & 71.8 & \textbf{92.6} & 27.5 & 40.0 & \textbf{94.7} & 62.4 & \textbf{97.9} & 81.5 & \textbf{50.0} \\
\midrule
\multicolumn{10}{c}{\textit{Strict budget}} \\
\midrule
gemini-3-flash-preview & 9.4 & 30.2 & 2.5 & 5.0 & 11.9 & 7.9 & 37.0 & 24.9 & 5.9 \\
qwen3-vl-plus          & 0.7 & 2.0  & 0.0 & 0.0 & 8.5  & 0.5 & 61.4 & 1.6  & 1.6 \\
gemma-4-31b-it         & 2.0 & 18.8 & 0.0 & 0.0 & 3.7  & 1.6 & 49.7 & 14.8 & 3.5 \\
kimi-k2.5              & 0.7 & 8.1  & 0.0 & 0.0 & 0.5  & 0.5 & 29.1 & 6.3  & 6.7 \\
gpt-5.2                & 1.3 & 9.4  & 0.0 & 0.0 & 1.1  & 1.1 & 40.7 & 7.4  & 0.0 \\
claude-sonnet-5        & \textbf{47.0} & \textbf{60.4} & \textbf{5.0} & \textbf{15.0} & \textbf{85.2} & \textbf{38.1} & \textbf{88.4} & \textbf{50.8} & \textbf{22.0} \\
\bottomrule
\end{tabularx}
\caption{Main results on the Core Structural Suite, denoted Level 1 with 149 levels; the Compositional Challenge Suite, denoted Level 2 with 40 levels; and all 189 levels. V@k and S@k denote Valid@k and Success@k within $k$ model calls. R@3 denotes recovery among repair-eligible physical failures. Loose budget relaxes material cost, whereas strict budget enforces the calibrated per-level limit. All entries are percentages; higher is better.}
\label{tab:main-results}
\end{table*}

\subsection{Experimental Setup}

We evaluate Gemini-3-Flash-Preview, Qwen3-VL-Plus, Gemma-4-31B-IT, Kimi-K2.5, GPT-5.2, and Claude-Sonnet-5 on all 189 levels. Level~1 denotes the 149 levels in the Core Structural Suite, and Level~2 denotes the 40 levels in the Compositional Challenge Suite. Every model receives semantically equivalent scene information, engineering constraints, and topology-output specifications and follows the common protocol for model calls, physics execution, and success detection described above.

We compare loose- and strict-budget settings. The strict setting enforces the material budget calibrated for each level from a manually constructed reference design as a hard deterministic constraint. The loose setting relaxes the material-cost restriction so that topology generation and physical execution can be examined when budget is not the primary bottleneck. Both settings allow at most three LLM calls and three physics attempts per episode. An invalid design consumes one model call but does not enter simulation; only a validated design is written into the game and executed.

We report results for Level~1, Level~2, and Overall. For each suite, we report Success@1 and Success@3; for Overall, we additionally report Valid@1, Valid@3, and Recovery@3. Strict budget is the primary setting for assessing executable design under constraints. Loose budget is a diagnostic condition that separates material-budget failures from limitations in topology generation and dynamic load bearing. All table entries are percentages, and Recovery@3 uses the eligible episodes defined in the evaluation protocol as its conditional denominator.

\subsection{Main Results}

Table~\ref{tab:main-results} reports performance under both budget settings. The gap between Valid and Success measures how often deterministic compliance fails to yield dynamic functionality. The change from @1 to @3 measures the benefit of bounded interaction. Strict budget is the primary benchmark condition; loose budget is a diagnostic setting that removes material cost as the dominant bottleneck.

\begin{figure*}[t]
    \centering
    \includegraphics[width=\textwidth]{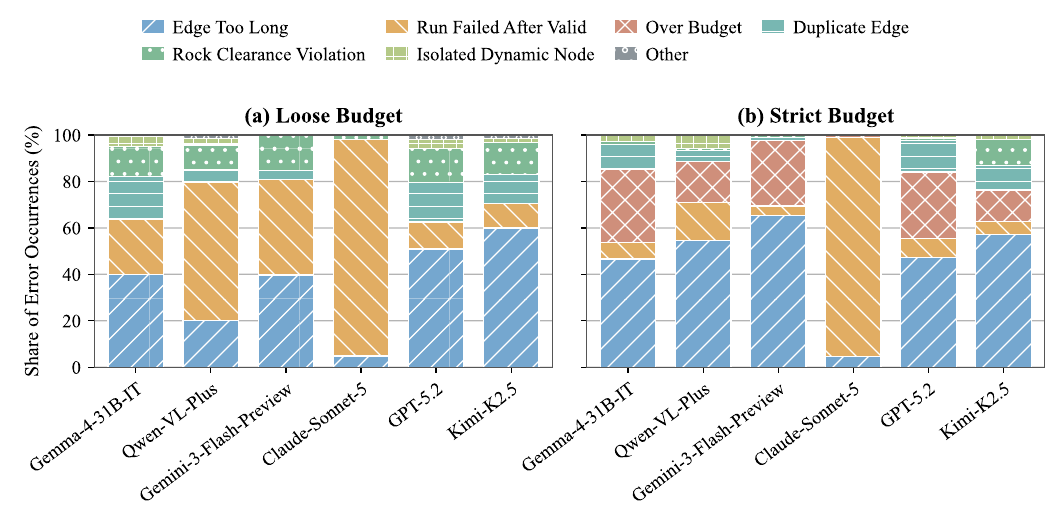}
    \caption{Composition of recorded error events over all three interaction rounds under loose and strict budgets. Each bar is normalized separately within a model and budget setting, so segment height is a within-model event share rather than a level failure rate or an absolute error count.}
    \label{fig:error-distribution}
\end{figure*}

\begin{figure*}[t]
    \centering
    \includegraphics[width=\textwidth]{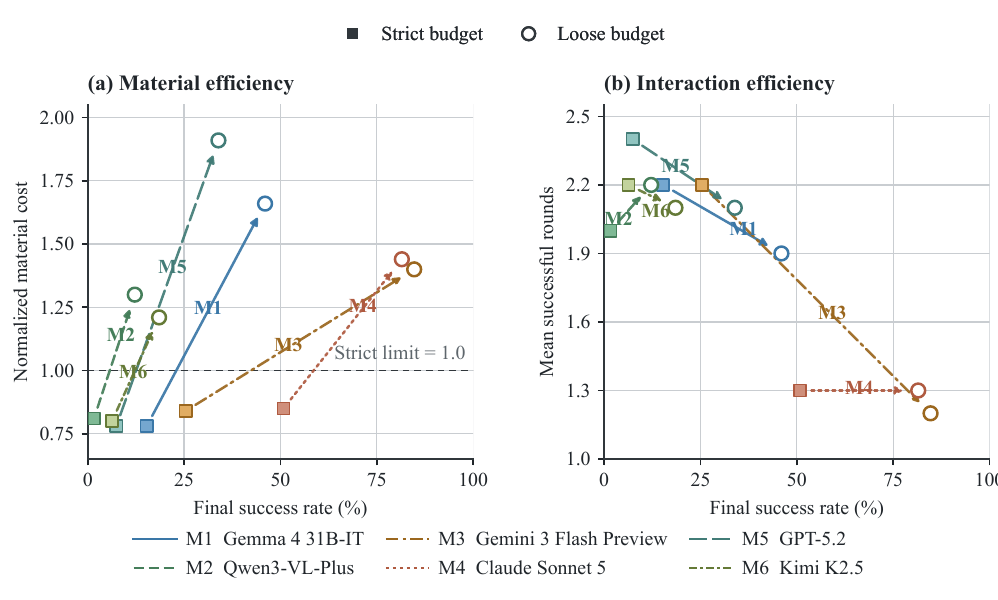}
    \caption{Success--efficiency trade-offs under loose and strict budgets. Colors identify models, squares denote strict budget, hollow circles denote loose budget, and arrows point from strict to loose. The horizontal axis is Overall Success@3 across all 189 levels. The left panel reports successful-design cost normalized by the corresponding per-level strict limit; its shaded region is budget compliant. The right panel reports model calls averaged over successful episodes. Both vertical axes are conditioned on success.}
    \label{fig:efficiency-tradeoff}
\end{figure*}

Recovery@3 has a conditional denominator and should not be read as another aggregate success rate. It measures recovery only among levels that experienced a physical failure while retaining sufficient budget for another model call and physics attempt.

\paragraph{Strict budgets change model rankings.}
Under loose budget, Gemini-3-Flash-Preview obtains the highest Overall Success@3 at 84.7\%. Under strict budget, Claude-Sonnet-5 ranks first with 50.8\%. High success under the loose setting therefore does not establish that a model can generate executable structures under constrained engineering conditions. The per-level material budget is part of the task definition, not an auxiliary cost measure.

\paragraph{Deterministic validity is not a reliable proxy for dynamic functionality.}
Under loose budget, Qwen3-VL-Plus reaches 92.1\% Valid@3 but only 12.2\% Success@3. A similar gap remains under strict budget. Models can use multiple rounds to produce topologies that satisfy explicit rules without ensuring vehicle traversal under native dynamics. Validator pass rates and physical success must therefore be reported separately.

\paragraph{Compositional constraints amplify the functional bottleneck.}
Level~2 is more difficult for every model. Under strict budget, only Gemini-3-Flash-Preview and Claude-Sonnet-5 achieve nonzero Success@3 on this subset, and the best score is 15.0\%. Combining material restrictions, clearance requirements, platforms, loads, and support conditions sharply reduces the feasible design space. Models that solve some core structural levels do not retain the same functional performance when several constraints apply simultaneously.

\paragraph{Multi-round gains are not equivalent to recovery from physical failure.}
Success@3 exceeds Success@1 for every model under both budget settings, whereas the highest strict-budget Recovery@3 is only 22.0\%. The increase in Success@3 also includes legality corrections prompted by validator feedback and cannot be attributed entirely to effective repair after execution failure. Success@3 and Recovery@3 must therefore be interpreted jointly rather than treating every multi-round gain as physical repair.

\subsection{Where Does the Capability Break Down?}

The main results establish an aggregate gap between deterministic legality and dynamic functionality but do not identify the errors that constitute this gap. Figure~\ref{fig:error-distribution} aggregates error events over all three interaction rounds and normalizes them within each model. Deterministic errors cover member length, budget, duplicate edges, clearance, and isolated nodes. \texttt{RUN\_FALSE\_AFTER\_VALID} denotes a candidate that passed validation but did not succeed during physics execution. Because one candidate may trigger several validator errors, the figure describes the composition of recorded error events rather than the level failure rate; absolute model performance should still be read from Table~\ref{tab:main-results}.

The error composition already separates two model behaviors. Claude-Sonnet-5 is dominated by post-validation physics failures in both settings, whereas strict budget shifts most other models toward deterministic errors that prevent simulation.

Claude-Sonnet-5 fails primarily during dynamic execution. \texttt{RUN\_FALSE\_AFTER\_VALID} accounts for 93.2\% and 94.3\% of its recorded errors under loose and strict budgets, respectively, while deterministic errors are rare. This pattern agrees with its high Valid@1 and Valid@3: Claude usually produces complete topologies that satisfy explicit constraints, but converting a legal structure into one that carries the vehicle across remains the bottleneck. Qwen3-VL-Plus shows the same tendency under loose budget, explaining why its Valid@3 is high while its Success@3 remains low.

GPT-5.2 and Kimi-K2.5 more often violate local geometric constraints before reaching physics execution. Under loose budget, \texttt{EDGE\_TOO\_LONG} accounts for 50.9\% and 60.0\% of their error events, respectively, and remains the largest category under strict budget. Gemma-4-31B-IT also exhibits substantial member-length errors together with more duplicate edges. The validator can detect these failures without dynamic inference. Their primary difficulty is therefore converting coordinates, material-specific length limits, and connectivity requirements into a legal graph topology.

Strict budget shifts errors toward deterministic validation. For Gemini-3-Flash-Preview, \texttt{EDGE\_TOO\_LONG} and \texttt{OVER\_BUDGET} together account for 93.5\% of recorded errors, whereas \texttt{RUN\_FALSE\_AFTER\_VALID} accounts for only 4.2\%. Gemma-4-31B-IT and GPT-5.2 also produce substantial over-budget errors. The reduced share of physics failures does not indicate better dynamic reasoning; more candidates are blocked by length or cost constraints before they enter the game. This selection effect helps explain why strict budget lowers both Valid@1 and Success@3.

The bottleneck is therefore model dependent. Some models are limited first by representation and constraint satisfaction, whereas others reliably enter simulation but fail to produce functional structures. Validator feedback identifies the former errors directly, but \texttt{RUN\_FALSE\_AFTER\_VALID} reveals only that execution was unsuccessful, not how the failure unfolded. This decomposition localizes failures without isolating the causal value of any feedback component; temporal frames remain part of the standard post-failure package rather than a separately varied intervention.

\subsection{How Many Resources Does Success Require?}

Figure~\ref{fig:efficiency-tradeoff} analyzes the efficiency of successful designs along material and interaction dimensions. The horizontal axis is Overall Success@3 across all 189 levels. Material efficiency is the cost of a successful design divided by the strict budget for that level and then averaged over successful levels. Interaction efficiency is the mean number of LLM calls among successful episodes. Squares and hollow circles distinguish strict and loose budgets, and arrows connect the two settings for the same model. Both vertical axes are conditioned on success, so failed levels and levels that never enter physics execution are not assigned zero cost.

A lower-right position combines broader benchmark success with lower resource use among successful episodes. This is a conditional trade-off, not an unconditional cost ranking, because models can succeed on subsets of different size and difficulty.

More material does not guarantee functional success. Under loose budget, Gemini-3-Flash-Preview achieves 84.7\% Success@3 with a mean cost ratio of 1.40. GPT-5.2 uses more material but succeeds much less often. Differences between models cannot be explained by material expenditure alone; structural organization still determines whether a design survives dynamic execution.

Similar budget utilization can correspond to markedly different functional performance. Under strict budget, the mean cost ratios of Claude-Sonnet-5 and Gemini-3-Flash-Preview differ by only 0.01, but their Success@3 scores differ by 25.4 percentage points. Budget compliance is necessary for an admissible design, but it does not remove differences in structural planning and dynamic load-bearing ability.

Interaction rounds indicate how strongly successful trajectories depend on later revision. Successful Claude-Sonnet-5 episodes use 1.3 rounds on average under both budget settings, suggesting that its strict-budget successes do not primarily depend on repeated repair. Gemini-3-Flash-Preview requires one additional LLM call on average under strict budget than under loose budget, indicating greater reliance on later revisions. Other models show related tendencies, but differences in the size and difficulty of their successful subsets prevent attributing round-count differences solely to the budget constraint.

All-episode resource use remains necessary for a complete efficiency comparison because a model with low success may solve only easier levels and obtain a low success-conditioned cost. Model calls and physics attempts over all episodes must therefore be reported separately. Overall, neither additional material nor additional repair rounds consistently corresponds to higher physical success.

\section{Conclusion}

PolyBridgeBench tests whether MLLMs transform visual scenes and engineering constraints into bridge topologies that survive native dynamic execution. Deterministic checks and simulation distinguish rule compliance from physical functionality, while fixed budgets quantify repair. Valid@k, Success@k, and Recovery@k separately measure validity, functionality, and post-failure recovery. It provides a reproducible, auditable test of structures that function physically rather than appear plausible or satisfy explicit rules.

\bibliography{aaai2027}



\appendix
\setcounter{secnumdepth}{2}

\raggedbottom
\setcounter{topnumber}{3}
\setcounter{dbltopnumber}{2}
\renewcommand{\topfraction}{0.95}
\renewcommand{\dbltopfraction}{0.95}
\renewcommand{\textfraction}{0.05}
\renewcommand{\floatpagefraction}{0.80}
\renewcommand{\dblfloatpagefraction}{0.80}
\makeatletter
\setlength{\@fptop}{0pt}
\setlength{\@fpsep}{10pt plus 2pt minus 2pt}
\setlength{\@fpbot}{0pt plus 1fil}
\setlength{\@dblfptop}{0pt}
\setlength{\@dblfpsep}{12pt plus 2pt minus 2pt}
\setlength{\@dblfpbot}{0pt plus 1fil}
\makeatother

\section{Benchmark Construction and Composition}
\label{app:benchmark-construction}

This section documents how the 189 executable levels are constructed, which
information is retained for analysis, and how the strict and loose material
budgets are defined.  The benchmark contains two complementary evaluation
strata.  The \emph{Core Structural Suite} contains 149 levels from 21
parameterized structural families.  The \emph{Compositional Challenge Suite}
contains 40 levels from six families that combine clearance, material, load,
vehicle, platform, and support constraints.  These strata are not an
in-distribution/out-of-distribution split: every evaluated model is run on
every level.

\subsection{Generation and Export Pipeline}
\label{app:generation-pipeline}

Each family is defined by a deterministic generator and a finite parameter
grid.  The grid varies bridge span, bank elevation, anchor layout, center
terrain, admissible materials, vehicle count, and vehicle weight.  Once a
level specification is fixed, the generator exports three synchronized
artifacts:
\begin{itemize}
    \item \texttt{<level>.json}, the complete native level layout, including
    terrain, vehicles, targets, resources, and the encoded initial save;
    \item \texttt{<level>\_inner.json}, the decoded initial editable topology
    containing the native kinematic anchors; and
    \item \texttt{<level>\_context.json}, the task-facing context and the
    analysis-only metadata described in Section~\ref{app:information-boundary}.
\end{itemize}
The three files share the same stable level name.  The controller refuses to
load an episode if any member of this triplet is absent or if the context does
not contain materials, budget, and vehicle-weight information.

\subsection{Core Structural Suite}
\label{app:core-suite}

Tables~\ref{tab:app-core-families-1} and
\ref{tab:app-core-families-2} enumerate the 21 mutually exclusive Core
families.  The family label is used only for aggregation.  Within a family,
levels vary one or more continuous or discrete parameters while preserving
the stated structural motif.  ``RW'' denotes a Road/Wood-only material set;
all other rows expose Road, Wood, and Steel.  Vehicle weight is expressed in
the game's passenger-car weight unit, denoted PG in the released context.

\begin{table*}[t]
\centering
\scriptsize
\setlength{\tabcolsep}{3.5pt}
\begin{tabularx}{\textwidth}{@{}l r c X X c@{}}
\toprule
\textbf{Family} & \textbf{$N$} & \textbf{Span (m)} &
\textbf{Primary structural variation} & \textbf{Material/load variation} &
\textbf{Strict budget} \\
\midrule
T1B & 5  & 8--16  & Standard flat span with two bank anchors
     & One vehicle; two vehicles for longer spans & \$7.0k--\$17.0k \\
T1C & 4  & 8--14  & Left-asymmetric bank support with three anchors
     & Full material set; one vehicle & \$7.0k--\$13.0k \\
T1D & 5  & 8--16  & Symmetric dual bank supports with four anchors
     & RW only; one vehicle & \$7.0k--\$17.0k \\
T1E & 15 & 8--16  & One centered bottom anchor at offsets 2, 3, or 4 m
     & RW at 2 m offset, otherwise full materials & \$7.0k--\$17.0k \\
T1F & 10 & 10--18 & Two bottom anchors at offsets 2 or 3 m
     & RW at 2 m offset, otherwise full materials & \$8.5k--\$17.5k \\
T1G & 16 & 10--16 & One off-center bottom anchor at 0.35 or 0.65 of span
     & Bottom offsets 2.5 or 3.5 m; one vehicle & \$8.5k--\$17.0k \\
T1H & 4  & 8--14  & Right-asymmetric mirror of T1C
     & Full material set; one vehicle & \$7.0k--\$13.0k \\
\bottomrule
\end{tabularx}
\caption{Core Structural Suite, part I: basic connectivity and anchor-layout
families.  Budget ranges are the per-level limits used in the strict-budget
condition.}
\label{tab:app-core-families-1}
\end{table*}

\begin{table*}[t]
\centering
\scriptsize
\setlength{\tabcolsep}{3.5pt}
\begin{tabularx}{\textwidth}{@{}l r c X X c@{}}
\toprule
\textbf{Family} & \textbf{$N$} & \textbf{Span (m)} &
\textbf{Primary structural variation} & \textbf{Material/load variation} &
\textbf{Strict budget} \\
\midrule
T2A & 8  & 8--14  & Uphill banks with 2 or 4 m elevation change
     & Two vehicles on the longer spans & \$7.0k--\$15.0k \\
T2B & 12 & 14--18 & Raised center platform with varied width and height
     & RW or full materials & \$7.5k--\$11.0k \\
T2C & 4  & 12--18 & Central rock-clearance constraint
     & Full materials; one vehicle & \$10.0k--\$15.0k \\
T2D & 8  & 8--14  & Downhill banks with 2 or 4 m elevation change
     & Two vehicles on the longer spans & \$7.0k--\$15.0k \\
T2E & 6  & 10--14 & Uphill banks plus a center-bottom anchor
     & Elevation change 2 or 4 m & \$8.0k--\$15.0k \\
T2F & 6  & 10--14 & Downhill banks plus a center-bottom anchor
     & Elevation change 2 or 4 m & \$8.0k--\$15.0k \\
T2G & 12 & 14--18 & Lowered center platform with varied width and depth
     & RW or full materials & \$6.5k--\$10.5k \\
T2H & 2  & 16--18 & Raised center platform; left bank above right bank
     & Full materials; one vehicle & \$9.5k--\$10.0k \\
T2I & 2  & 16--18 & Lowered center platform; left bank above right bank
     & Full materials; one vehicle & \$9.5k--\$10.0k \\
T2J & 2  & 16--18 & Raised center platform; right bank above left bank
     & Full materials; one vehicle & \$9.0k--\$10.0k \\
T2K & 2  & 16--18 & Lowered center platform; right bank above left bank
     & Full materials; one vehicle & \$9.0k--\$10.0k \\
T4B & 10 & 8--16  & Flat spans under elevated vehicle load
     & Weight 3 or 5 PG; one or two vehicles & \$7.0k--\$28.0k \\
T4C & 8  & 10--16 & Uphill/downhill spans under elevated load
     & Two vehicles, each 3 PG & \$11.0k--\$42.0k \\
T4D & 8  & 10--16 & Elevated load plus a center-bottom anchor
     & Three 3-PG vehicles or one 5-PG vehicle & \$10.0k--\$20.0k \\
\bottomrule
\end{tabularx}
\caption{Core Structural Suite, part II: terrain, clearance, and dynamic-load
families.}
\label{tab:app-core-families-2}
\end{table*}

\subsection{Compositional Challenge Suite}
\label{app:compositional-suite}

The six Compositional families deliberately combine constraints that appear
separately in the Core suite.  They are therefore reported as a second
evaluation stratum rather than as evidence of training-time distribution
shift.  Table~\ref{tab:app-compositional-families} lists their composition.

\begin{table*}[t]
\centering
\scriptsize
\setlength{\tabcolsep}{3.5pt}
\begin{tabularx}{\textwidth}{@{}l r c X X c@{}}
\toprule
\textbf{Family} & \textbf{$N$} & \textbf{Span (m)} &
\textbf{Constraint composition} & \textbf{Material/load variation} &
\textbf{Strict budget} \\
\midrule
G1 & 8 & 12--18 & Rock clearance combined with increased traffic load
   & Full materials; two standard or one 3-PG vehicle & \$16.0k--\$21.0k \\
G2 & 4 & 12--14 & Rock clearance, restricted materials, and increased load
   & RW only; two standard or one 3-PG vehicle & \$11.0k--\$12.0k \\
G3 & 6 & 14--18 & Raised center platform and off-center rock clearance
   & Full materials; rock placed on either side & \$10.0k--\$13.0k \\
G4 & 6 & 14--18 & Lowered center platform and off-center rock clearance
   & Full materials; rock placed on either side & \$10.0k--\$17.0k \\
G5 & 8 & 16--18 & Raised/lowered platform, clearance, and increased load
   & Two standard or one 3-PG vehicle & \$12.0k--\$13.0k \\
G6 & 8 & 12--24 & Asymmetric or off-center support combined with clearance
   & Full materials; one vehicle & \$16.0k--\$24.0k \\
\bottomrule
\end{tabularx}
\caption{The six families in the 40-level Compositional Challenge Suite.}
\label{tab:app-compositional-families}
\end{table*}

\subsection{Strict and Loose Material Budgets}
\label{app:budget-calibration}

Material cost is computed from member length rather than from member count.
For a generated topology $G$ with members $m$, the controller computes
\begin{equation}
    \operatorname{Cost}(G)=\sum_{m\in G}c_{\tau(m)}
    \left\lVert p_{a(m)}-p_{b(m)}\right\rVert_2,
\end{equation}
where $\tau(m)$ is the material type, $c_{\tau(m)}$ is its per-meter cost,
and $a(m)$ and $b(m)$ are the member endpoints.  Costs are accumulated before
rounding the reported total to two decimal places.

The strict limit $B_i$ for each level was calibrated by manually loading that
level into Poly Bridge, constructing and executing a successful reference
bridge, and recording a level-specific admissible budget.  Calibration was
performed before model evaluation.  Only the scalar limit is exposed to a
model; the reference topology is not part of the model input and is not used
as an evaluation target.  The resulting 189 limits contain 26 distinct values
and range from \$6,500 to \$42,000.  The strict condition rejects a candidate
when $\operatorname{Cost}(G)>B_i$.  The loose condition preserves the same
level geometry and task constraints but uses a uniform \$50,000 limit, making
material cost unlikely to be the first gating failure.  The loose condition
is diagnostic and does not replace the strict condition as the primary
benchmark setting.

\subsection{Clearance-Region Calibration}
\label{app:clearance-calibration}

The rendered rock sprite does not expose metric collision geometry.  We
therefore define a deterministic triangular forbidden region from the stored
sprite position $(x_r,y_b)$ and scale $(w,h)$.  Its vertices are
\begin{equation}
  (x_r+w,\;y_b+3.5h),\qquad
  (x_r-6w,\;y_b),\qquad
  (x_r+6w,\;y_b).
\end{equation}
For the benchmark rock scale $(w,h)=(0.5,2.0)$, this gives a base extending
3 m to either side and an apex 7 m above the stored bottom coordinate, shifted
0.5 m in $x$.  The mapping is fixed before evaluation and is identical for all
models.  The polygon, rather than the raw sprite scale, is supplied in
\texttt{semantic\_environment.forbidden\_regions} and used by the deterministic
clearance validator.

\subsection{Information Boundary and Analysis Metadata}
\label{app:information-boundary}

Table~\ref{tab:app-context-boundary} distinguishes task information from
offline annotations.  The controller constructs the model-facing level
context from exactly two objects: \texttt{task\_context} and
\texttt{semantic\_environment}.  The context retains \texttt{levelName} as an
episode identifier, but it does not append the separate
\texttt{analysis\_metadata.family\_id}, generator parameter dictionary, or
level tags.  Family-level and tag-level breakdowns are computed only after an
episode is complete.

\begin{table*}[t]
\centering
\small
\setlength{\tabcolsep}{4pt}
\begin{tabularx}{\textwidth}{@{}l X X@{}}
\toprule
\textbf{Scope} & \textbf{Fields} & \textbf{Role} \\
\midrule
Model-visible task context &
\texttt{levelName}, \texttt{available\_materials}, \texttt{budget\_limit},
\texttt{vehicle\_count}, \texttt{vehicle\_weights\_pg},
\texttt{vehicle\_weight\_pg}, \texttt{material\_profile} &
Defines the admissible materials, load, and cost constraint. \\
Model-visible semantic environment &
\texttt{terrain\_affordances}, \texttt{forbidden\_regions} &
Defines drivable center terrain and deterministic clearance polygons. \\
Model-visible save state &
Native node IDs, coordinates, and \texttt{isKinematic} values &
Defines the immutable structural attachment points and current topology. \\
Analysis only &
\texttt{family\_id}, \texttt{generator\_tags},
\texttt{generator\_params}, \texttt{level\_tags} &
Used for aggregation and diagnostic breakdowns; not appended to the
model-facing context. \\
Generation audit only &
Raw terrain, scenery, vehicle, target, and resource parameters &
Retained to regenerate and inspect a level; not provided as prompt context. \\
\bottomrule
\end{tabularx}
\caption{Boundary between model-visible task information and metadata retained
for generation or offline analysis.}
\label{tab:app-context-boundary}
\end{table*}

\section{Task Interface and Structured Representation}
\label{app:task-interface}

\subsection{Observation and Action Contract}
\label{app:observation-action}

At the first call, the model receives (i) an editor screenshot, (ii) the
model-visible level context from Table~\ref{tab:app-context-boundary}, and
(iii) the current decoded save containing the native anchors.  At a later
call, the same fields are augmented with the previous candidate topology and
the most recent deterministic or physical feedback.  After a validator
failure, the exact invalid draft and validator message are returned.  After a
physical failure, the current topology, a conservative
\texttt{failure\_or\_timeout} summary, and selected temporal frames are
returned.  Each call must regenerate a complete topology; the model cannot
issue an imperative action such as ``add one beam'' without returning the
resulting full graph.

The primary benchmark uses no manually authored or trajectory-derived skill
instructions.  Thus, the optional skill field in the controller is empty in
the reported evaluation.

\subsection{Node--Member--Material Graph}
\label{app:graph-schema}

A candidate bridge is a graph $G=(V,M)$.  Every node $v\in V$ contains an
integer identifier, a two-dimensional position $(x,y)$ in game-world meters,
and an \texttt{isKinematic} flag.  Native anchors are kinematic and must retain
their original identifiers and coordinates.  Model-created nodes are dynamic
and must set \texttt{isKinematic} to false.  Every member $m\in M$ contains an
integer identifier, a material type, the identifiers of its two endpoint
nodes, and a \texttt{rate} field retained for save-format compatibility.

The benchmark levels reported in the paper expose the three materials in
Table~\ref{tab:app-materials}.  A particular level may restrict this set to
Road and Wood.  Material strength values are supplied as design context, but
the controller does not analytically predict capacity from them; native game
execution remains the only functional test.

\begin{table}[!h]
\centering
\small
\begin{tabular}{@{}l c c c@{}}
\toprule
\textbf{Material} & \textbf{Type} & \textbf{Max. length} &
\textbf{Cost/m} \\
\midrule
Road  & 1 & 2.0 m & \$200 \\
Wood  & 2 & 2.0 m & \$180 \\
Steel & 3 & 4.0 m & \$450 \\
\bottomrule
\end{tabular}
\caption{Material identifiers and deterministic limits in the released
189-level benchmark.}
\label{tab:app-materials}
\end{table}

The required model output is one raw JSON object with the following shape.
The \texttt{thought\_process} field is a brief decision summary used for
trajectory audit; it is not treated as privileged model reasoning and is not
scored.

\begingroup
\small
\begin{verbatim}
{
  "action": "modify",
  "thought_process":
    "brief decision summary",
  "nodes": [
    {"id": 1, "x": 0.0, "y": 0.0,
     "isKinematic": true},
    {"id": 2, "x": 2.0, "y": 0.0,
     "isKinematic": false}
  ],
  "edges": [
    {"id": 3, "type": 1,
     "anchorAID": 1, "anchorBID": 2,
     "rate": 0}
  ]
}
\end{verbatim}
\endgroup

Node coordinates are rounded to four decimal places during conversion.  A
node becomes a save object with \texttt{type=0}.  A member becomes a save
object whose \texttt{type} is the material identifier and whose stored
$(x,y)$ coordinate is the midpoint of its endpoints.  The endpoint IDs remain
in \texttt{anchorAID} and \texttt{anchorBID}; consequently, the graph can be
validated before it is serialized into the native save format.

\subsection{Model-Visible Context Schema}
\label{app:model-context-schema}

The following abridged example illustrates the exact nesting of the
model-facing context.  Fields with empty lists remain present so that all
models receive the same schema.

\begingroup
\small
\begin{verbatim}
{
  "task_context": {
    "levelName": "<episode identifier>",
    "available_materials":
      ["Road", "Wood", "Steel"],
    "budget_limit": 13000,
    "vehicle_count": 2,
    "vehicle_weights_pg": [3.0, 3.0],
    "vehicle_weight_pg": 3.0,
    "material_profile": "full"
  },
  "semantic_environment": {
    "terrain_affordances": [
      {
        "type": "center_platform",
        "shape": "rectangular_terrain",
        "x_start": 6.0,
        "x_end": 10.0,
        "surface_y": 2.0,
        "bottom_y": -5.0,
        "drivable": true,
        "support_anchor_ids": [3, 4]
      }
    ],
    "forbidden_regions": [
      {
        "type": "rock",
        "shape": "triangle",
        "vertices": [
          {"x": 5.0, "y": 2.0},
          {"x": 1.5, "y": -5.0},
          {"x": 7.5, "y": -5.0}
        ],
        "applies_to":
          ["Road", "Wood",
           "Steel", "Hydraulics"]
      }
    ]
  }
}
\end{verbatim}
\endgroup

The structured context is authoritative for task constraints.  The image is
used to ground the visible scene and current editor state, while the decoded
save is authoritative for exact node identifiers and coordinates.  When
sources appear inconsistent, the prompt instructs the model to follow the
structured constraint and save representations rather than infer hidden
generator parameters.

\subsection{Prompt Specification}
\label{app:prompt-specification}

The static instruction below is a line-wrapped transcription of the prompt
used for the primary benchmark.  Line wrapping is changed for typesetting;
the wording and operative constraints are preserved.  The concrete JSON
example embedded in the runtime prompt follows the schema already shown in
Section~\ref{app:graph-schema}.

\begingroup
\scriptsize
\begin{verbatim}
# Role
You are an expert Structural Engineer and Level
Designer
for Poly Bridge. Generate or repair a complete bridge
topology from image + save code + level context.

Use engineering heuristics with material stats/costs;
native game execution alone determines physical
success.
Some levels may contain 2 or 3 vehicles. Read
task_context.vehicle_count and
task_context.vehicle_weights_pg; design for every
vehicle.

Level context is intentionally scoped to task_context
and
semantic_environment; raw generator parameters are not
part of the task.

If semantic_environment.terrain_affordances includes
center platforms/islands, use the screenshot and
kinematic anchor
positions to distinguish unsupported gaps from solid
drivable platform terrain. The vehicle path must be
continuous, but road edges are only necessary over
unsupported gaps unless a structural reason requires
them.

If semantic_environment.forbidden_regions is
non-empty,
keep Road/Wood/Steel/Hydraulics geometry outside those
regions. Treat forbidden-region vertices as the
geometry
used by the controller validator.

When temporal physics frames are provided, compare
their
progression and use observed deformation, collapse, or
vehicle obstruction to guide repair; treat causal
explanations as hypotheses. failure_or_timeout means
the controller did not confirm success within its
fixed window;
it does not prove an exact failure cause.

# Material Encyclopedia (for reasoning)
WARNING: You cannot use all materials every level.
Only materials in task_context.available_materials are
legal.
- Road (Type 1): strength 900PN, max length 2m,
  cost $200/m.
- Wood (Type 2): strength 800PN, max length 2m,
  cost $180/m.
- Steel (Type 3): strength 2000PN, max length 4m,
  cost $450/m.

# Generation Protocols (Strict Rules)
1. Nodes and edges
- Preserve every original isKinematic=true base node
  from
  Current Save Code with the same id and coordinates.
- New suspended nodes must set isKinematic=false.
  Never
  invent a new isKinematic=true node.
- IDs for newly added nodes must continue
  sequentially from
  the highest existing ID in Current Save Code.
- Edge type is the material identifier. Never use
  type 0
  for an edge; type 0 is reserved for nodes.

2. Strict length limits
- Calculate Euclidean endpoint distance.
- Road/Wood <= 2.0m; Steel <= 4.0m.
- If a limit is exceeded, insert an intermediate node.
- Do not duplicate the same material between the same
  two
  nodes; repeated identical edges are invalid.

3. Evidence roles and constraints
- task_context and semantic_environment are
  authoritative
  for structured constraints.
- Save Code is authoritative for current topology and
  exact
  node geometry.
- Validation Feedback is authoritative for
  deterministic
  violations in the previous draft.
- Use editor images for scene/current-state grounding
  and
  temporal frames as execution evidence for repair.
- Treat visual causal inferences as hypotheses and
  discard
  conflicting History.
- Use only materials in
  task_context.available_materials.
- If task_context.budget_limit is provided, keep
  estimated
  material cost <= that limit.
- If validation or physics feedback lists errors,
  repair
  them in the returned topology.
- The controller automatically validates and runs
  every
  legal topology.

# Output Format (STRICT)
Return one valid raw JSON object only: no Markdown
block or extra text. Always return action="modify".
The required
fields are thought_process, nodes, and edges.

thought_process is a brief decision summary, not a
step-by-step derivation. For an initial generation,
state the structural concept and how loads reach the
original anchors; if forbidden regions exist, state
the clearance route. For repair, state the observed
validation or physics evidence, mark uncertain
physical causes as hypotheses, and
name the concrete topology changes.
\end{verbatim}
\endgroup

For every call, the user message inserts the current
\texttt{task\_context}/\texttt{semantic\_environment} JSON and decoded save.
Depending on the previous outcome, it then appends one of the following:
\begin{itemize}
    \item the exact validator message and invalid draft topology;
    \item a compact physical-execution summary and the selected temporal
    evidence; or
    \item neither, for the initial call.
\end{itemize}
The message ends with labeled visual inputs.  The label
\texttt{editor\_static} identifies the current editable state;
\texttt{physics\_t\_<time>} identifies an individual frame; and
\texttt{physics\_contact\_sheet\_<index>} identifies a timestamped frame
grid.  In the primary setting, four uniformly sampled frames are packaged as
one contact sheet.

\section{Deterministic Validation and Native Execution}
\label{app:validation-execution}

\subsection{Deterministic Legality Checks}
\label{app:validator-rules}

The validator defines the legal action space before native execution.  It is
not a bridge generator, structural optimizer, or physics oracle.  It can
reject a graph for a factual schema, geometry, connectivity, material,
clearance, or cost violation, but it does not estimate load capacity,
stability, deformation, or successful vehicle traversal.

Table~\ref{tab:app-validator-rules} lists the hard checks.  All applicable
violations found in a candidate are returned together.  A rejected candidate
consumes its LLM call, is never written into the live game save, and consumes
no physics attempt.

\begin{table*}[t]
\centering
\scriptsize
\setlength{\tabcolsep}{3.5pt}
\begin{tabularx}{\textwidth}{@{}l X X@{}}
\toprule
\textbf{Check} & \textbf{Acceptance condition} & \textbf{Purpose} \\
\midrule
Parse and object schema &
The response resolves to a \texttt{modify} action with complete node and edge
lists that can be converted into an \texttt{Objects} array. &
Prevents malformed or non-executable outputs. \\
Native anchors &
Every original kinematic node is present with the same ID and coordinates
within a $10^{-3}$ tolerance. &
Prevents removal or movement of level-provided supports. \\
Kinematic status &
A model-created node cannot set \texttt{isKinematic=true}. &
Prevents invention of unsupported fixed points. \\
Endpoint references &
Both endpoint IDs of every member occur in the node set. &
Ensures that every member defines a graph edge. \\
Material availability &
Each member type is included in the level's
\texttt{available\_materials}. &
Enforces level-specific material restrictions. \\
Stored midpoint &
The member object's stored coordinate equals the endpoint midpoint within
$10^{-3}$. &
Maintains consistency with the native save representation. \\
Member length &
Road and Wood are at most 2.0 m; Steel and Hydraulics are at most 4.0 m. &
Enforces native per-segment construction limits. \\
Duplicate member &
For a fixed material type and unordered endpoint pair, at most one member may
appear. &
Prevents no-op duplication from being counted as reinforcement. \\
Dynamic-node degree &
Every model-created node has at least one incident material member. &
Rejects isolated nodes. \\
Anchor connectivity &
Every connected component containing a model-created node also contains at
least one native kinematic anchor. &
Rejects floating structural components. \\
Clearance &
No applicable dynamic node lies inside a forbidden polygon and no applicable
member segment intersects it. Boundary contact counts as intersection. &
Enforces the calibrated rock-clearance action space. \\
Material cost &
In strict mode, the summed length-weighted cost is at most the level-specific
budget. In loose mode the uniform limit is \$50,000. &
Makes admissibility under resource constraints explicit. \\
\bottomrule
\end{tabularx}
\caption{Deterministic checks applied before native physics execution.}
\label{tab:app-validator-rules}
\end{table*}

For connectivity, the implementation constructs a disjoint-set forest over
all nodes and unions the endpoints of every referenced material member.  A
dynamic node is valid only if its component root matches the root of at least
one native kinematic anchor.  Clearance uses standard point-in-polygon and
segment--polygon intersection tests; nodes or segments on the polygon boundary
are treated as violations.

\subsection{Feedback Taxonomy}
\label{app:error-taxonomy}

Validator messages are retained verbatim for the next repair call and are
also normalized into the audit codes in Table~\ref{tab:app-error-codes}.
Normalization is used for aggregate failure analysis; it does not replace the
original factual feedback shown to the model.  A candidate can contribute
multiple deterministic codes.  By contrast,
\texttt{RUN\_FALSE\_AFTER\_VALID} marks an unsuccessful native execution after
all deterministic checks passed and is not a validator violation.

\begin{table*}[t]
\centering
\small
\setlength{\tabcolsep}{4pt}
\begin{tabularx}{\textwidth}{@{}l X@{}}
\toprule
\textbf{Audit code} & \textbf{Meaning} \\
\midrule
\texttt{MISSING\_BASE\_NODE} & A native kinematic anchor is absent. \\
\texttt{MOVED\_KINEMATIC\_NODE} & A native anchor coordinate was changed. \\
\texttt{MATERIAL\_NOT\_ALLOWED} & A member uses a material excluded by the level. \\
\texttt{ROCK\_CLEARANCE\_VIOLATION} & A node or member intersects a forbidden polygon. \\
\texttt{EDGE\_TOO\_LONG} & A member exceeds its material-specific length limit. \\
\texttt{DUPLICATE\_EDGE} & The same material appears more than once on an endpoint pair. \\
\texttt{DISCONNECTED\_FROM\_BASE} & A dynamic component contains no native anchor. \\
\texttt{ISOLATED\_DYNAMIC\_NODE} & A model-created node has degree zero. \\
\texttt{OVER\_BUDGET} & Length-weighted material cost exceeds the active limit. \\
\texttt{RUN\_FALSE\_AFTER\_VALID} & Native execution did not reach confirmed success. \\
\bottomrule
\end{tabularx}
\caption{Normalized error taxonomy used in the diagnostic analysis.}
\label{tab:app-error-codes}
\end{table*}

\subsection{Save Serialization and Game Loading}
\label{app:serialization}

Once a candidate passes validation, the controller saves the decoded
\texttt{DisplayName}/\texttt{Objects} JSON for audit.  It then serializes the
compact JSON to UTF-8, applies zlib compression, and Base64-encodes the result
using the game's save representation.  The encoded payload replaces the first
entry of the target level's \texttt{saveFiles} list; all other level metadata
is preserved.  The game-loading helper then reloads this layout in the editor.
This conversion is deterministic and does not alter the graph selected by the
model.

\subsection{Native Physics Execution and Success Detection}
\label{app:physics-execution}

Table~\ref{tab:app-execution-config} gives the primary execution settings.
The controller records an editor-state frame immediately before starting the
simulation, presses the native run control, and samples the screen at 4 fps
for a fixed 12-second window.  The final screenshot is inspected at a fixed
interface pixel.  A green-channel value above the threshold corresponds to
the game's success state, which is reached only when the level's required
vehicle traversal completes.  Otherwise the controller records
\texttt{failure\_or\_timeout}.  This label deliberately makes no claim about
whether the bridge collapsed, blocked a vehicle, or merely failed to finish
within the observation window.

\begin{table}[t]
\centering
\small
\begin{tabular}{@{}l l@{}}
\toprule
\textbf{Setting} & \textbf{Primary value} \\
\midrule
Maximum LLM calls & 3 per episode \\
Maximum physics attempts & 3 per episode \\
Simulation window & 12 s \\
Raw capture rate & 4 fps \\
Success pixel & $(1095,513)$ \\
Success condition & green channel $>150$ \\
Feedback selection & uniform \\
Selected feedback frames & 4 \\
Feedback packaging & one contact sheet \\
Failure-marker pixel & $(1145,637)$ \\
Failure-marker condition & red channel $>150$ \\
Post-marker frames retained & 0 \\
\bottomrule
\end{tabular}
\caption{Controller and native-execution settings for the primary benchmark.}
\label{tab:app-execution-config}
\end{table}

\subsection{Temporal Evidence and Repair Loop}
\label{app:temporal-feedback}

For an unsuccessful execution, the controller first removes repeated frames
after the game displays its terminal failure marker.  It scans the captured
sequence chronologically and, if the red channel at pixel $(1145,637)$ exceeds
150, retains frames only through the first marked frame.  If no marker is
detected, the full 12-second sequence remains eligible; this preserves timeout
evidence rather than imposing an inferred failure time.

Four frames are then selected at uniformly spaced indices from the eligible
sequence, including its endpoints.  Each frame is annotated with its frame
index and elapsed time and placed in a single contact sheet.  The next model
call receives this sheet together with (i) a new editor screenshot after
returning from simulation, (ii) the full current topology, and (iii) a compact
summary containing the outcome, duration, detector, selected timestamps, and
cutoff metadata.  The prompt explicitly states that causal explanations based
on the frames are hypotheses.

The overall episode controller is summarized below.  LLM calls and physics
attempts have separate counters: validator rejection advances only the former,
whereas every valid candidate advances both before native execution.

\begingroup
\scriptsize
\begin{verbatim}
for call = 1, 2, 3:
    observation <- scene + context + current topology
                   + latest available feedback
    candidate <- MLLM(observation)

    if candidate cannot be parsed or converted:
        return deterministic feedback on next call
        continue

    violations, usage <-
        deterministic_validator(candidate)
    if violations are non-empty:
        return violations and invalid draft
        on next call
        continue

    write candidate to the native game save
    result, raw_frames <- execute for up to 12 seconds
    if result is success:
        terminate episode as successful

    if no call or physics budget remains:
        terminate episode as unsuccessful

    frames <- cutoff terminal repeats, then sample 4
    feedback <- failure_or_timeout summary
                + contact sheet
terminate when the call budget is exhausted
\end{verbatim}
\endgroup

For audit and replay, the controller retains the model response, parsed
topology, decoded design JSON, encoded save payload, validator feedback,
material-usage report, raw frames, selected-frame metadata, contact sheet,
per-round timing, and cumulative resource counters.  These records support
the layered validity, functionality, recovery, error, and efficiency analyses
without treating model-written explanations as ground-truth failure causes.

\section{Evaluation and Reproducibility Details}
\label{app:evaluation-reproducibility}

\subsection{Evaluation Unit and Interaction Accounting}
\label{app:evaluation-unit}

The atomic evaluation unit is one model--level--budget-condition episode.  A
paper-level result contains one episode for each of the 189 levels under the
specified model and budget condition.  The interaction horizon is indexed by
LLM calls rather than by physics attempts.  Call $t$ is the $t$-th complete
topology generation or repair request, regardless of whether its output
passes validation.  A validator rejection therefore advances $t$ but not the
physics-attempt counter.  A candidate that passes validation advances both
counters when it is executed.  In the released controller,
\texttt{rounds\_used} and \texttt{llm\_calls\_used} consequently have the same
value.

Each episode has at most three LLM calls and three physics attempts.  The two
limits are enforced independently.  For example, two validator-invalid
outputs followed by one valid execution consume three calls but only one
physics attempt.  Conversely, three valid but unsuccessful candidates consume
the full allowance of both resources.  An episode terminates immediately on
confirmed physical success or when no LLM call remains; the physics limit is
also checked after every execution.

\subsection{Metric Edge Cases}
\label{app:metric-edge-cases}

The main-paper definitions of Valid@$k$, Success@$k$, and Recovery@$k$ are
implemented directly from the round-aligned trajectory.  A round is valid
when it produces a candidate admitted to native execution.  A round is
successful only when that execution returns confirmed success.  Table
\ref{tab:app-metric-cases} makes the accounting explicit for common edge
cases.

\begin{table*}[t]
\centering
\small
\setlength{\tabcolsep}{4pt}
\begin{tabularx}{\textwidth}{@{}X c c c X@{}}
\toprule
\textbf{Round outcome} & \textbf{Consumes call} &
\textbf{Consumes physics} & \textbf{Valid at this call} &
\textbf{Success/recovery treatment} \\
\midrule
Unparseable response, wrong action, or missing required output field &
Yes & No & No & Cannot contribute to Success or Recovery. \\
Parsed graph rejected by deterministic validation &
Yes & No & No & May receive validator feedback on a later call. \\
Valid graph with \texttt{failure\_or\_timeout} &
Yes & Yes & Yes & Not successful; becomes recovery-eligible only when both
budgets permit a later execution. \\
Valid graph with confirmed success &
Yes & Yes & Yes & Contributes to Success@$k$ for every $k$ at or after this
call and terminates the episode. \\
No valid graph within three calls &
Up to 3 & 0 & No & Counts as failure for Success@3 and is not eligible for
Recovery@3. \\
\bottomrule
\end{tabularx}
\caption{Call-, physics-, validity-, and recovery-accounting rules.}
\label{tab:app-metric-cases}
\end{table*}

\paragraph{Valid@$k$.}
An episode contributes one if at least one candidate from calls
$1,\ldots,k$ enters physics.  Parsing and validator failures contribute zero.
The main results use $k\in\{1,3\}$; call-2 values can be derived from the same
trajectory for cumulative curves.

\paragraph{Success@$k$.}
An episode contributes one if a physics execution associated with a call no
later than $k$ succeeds.  Success@1 is therefore strict one-call end-to-end
success: the first response must be parseable, legal, and physically
functional.  It differs from \emph{first-physics-attempt success}.  If call 1
is validator-invalid and call 2 is the first candidate to enter physics and
succeed, first-physics-attempt success is true but Success@1 is false.

\paragraph{Recovery@3.}
A level is eligible only if it experiences an unsuccessful physical execution
before call 3 and before physics attempt 3, so that at least one call and one
physics attempt remain.  Recovery is computed per level, not per failure
event.  Let $F$ denote an unsuccessful physical execution, $S$ a successful
one, and $V$ a validator rejection.  Then $F\!\rightarrow\!S$ and
$F\!\rightarrow\!F\!\rightarrow\!S$ each count as one successful recovery;
$V\!\rightarrow\!S$ is not a physical recovery; and an $F$ on call 3 is not
eligible.  The denominator is model- and condition-dependent and must be
reported together with the numerator.  Recovery@3 is consequently a
conditional diagnostic, not another aggregate success rate.

\paragraph{Failure and error counts.}
Result statuses are mutually exclusive at the episode level, but validator
error codes are multi-label at the candidate level.  One rejected candidate
can therefore contribute both \texttt{EDGE\_TOO\_LONG} and
\texttt{OVER\_BUDGET}.  Normalized error-composition figures divide by the
number of recorded error events for that model and condition; they do not
estimate the fraction of failed levels.  Absolute event counts and the number
of levels reaching physics are retained alongside normalized shares.

\subsection{Resource and Efficiency Metrics}
\label{app:resource-metrics}

Resource use is reported with both an all-episode denominator and a
success-conditioned denominator.  For episode $i$, let $C_i$ be the number of
LLM calls, $A_i$ the number of physics attempts, $Y_i$ the final-success
indicator, $M_i$ the cost of its successful design when $Y_i=1$, and $B_i$ the
strict calibrated budget for that level.  We report
\begin{align}
\mathrm{AvgCalls}_{\mathrm{all}}
  &= \frac{1}{N}\sum_{i=1}^{N} C_i, \notag\\
\mathrm{AvgAttempts}_{\mathrm{all}}
  &= \frac{1}{N}\sum_{i=1}^{N} A_i,\\
\mathrm{AvgCalls}_{\mathrm{succ}}
  &= \frac{\sum_i Y_i C_i}{\sum_i Y_i}, \notag\\
\mathrm{AvgAttempts}_{\mathrm{succ}}
  &= \frac{\sum_i Y_i A_i}{\sum_i Y_i},\\
\mathrm{BudgetUtilization}_{\mathrm{succ}}
  &= \frac{1}{\sum_iY_i}
     \sum_{i:Y_i=1}\frac{M_i}{B_i}.
\end{align}
Failed episodes contribute their actual consumed calls and attempts to the
all-episode means.  In particular, an episode that never passes validation
contributes zero physics attempts rather than a missing value.  Material cost
is success-conditioned because an unsuccessful structure has no meaningful
functional-efficiency interpretation.  The successful subsets can differ in
size and difficulty across models; material-efficiency plots are therefore
interpreted jointly with Success@3 rather than as standalone rankings.  A
common-success subset may additionally be released for paired cost
comparisons, but it is not substituted for the full benchmark result.

Validation-repair calls count rounds ending in a parsing, geometry, or
deterministic validation failure.  Physics-repair calls count model requests
that receive feedback from a previous native execution.  These quantities are
separate because the former spend calls without entering physics, whereas the
latter spend both forms of test-time interaction if their repaired topology
is valid.

\subsection{Suite, Family, and Tag Aggregation}
\label{app:aggregation-protocol}

Overall metrics use all 189 levels.  Core metrics use the 149 non-G families,
and Compositional metrics use the 40 G-prefixed levels.  The stored
\texttt{family\_id} assigns each level to exactly one of the 27 families and
supports mutually exclusive family-level comparisons.  By contrast,
\texttt{level\_tags} such as \texttt{slope}, \texttt{multi\_vehicle},
\texttt{rock\_clearance}, and \texttt{road\_wood\_only} overlap.  Tag-level
breakdowns describe performance conditional on a constraint and must not be
summed to reconstruct suite totals.  Neither family IDs nor tags are used as
evaluation targets.

\subsection{Models and Request Controls}
\label{app:model-request-controls}

All models receive the task contract in Appendix~\ref{app:task-interface} and
the same image ordering.  Requests are made through an OpenAI-compatible
gateway with SDK-level automatic retries disabled; the experiment wrapper
records each explicit HTTP attempt.  Table~\ref{tab:app-model-controls}
documents the model aliases and request controls used by the current
benchmark harness.  ``Provider default'' means that no thinking intensity or
binary override is sent.  For Qwen and Kimi the gateway supports a binary
reasoning switch, but the primary setting leaves it unset.  Gemma exposes no
compatible reasoning control through this interface.

\begin{table*}[t]
\centering
\scriptsize
\setlength{\tabcolsep}{3.5pt}
\begin{tabularx}{\textwidth}{@{}l X c c c c@{}}
\toprule
\textbf{Reported model} & \textbf{API model identifier} &
\textbf{Reasoning request} & \textbf{Temperature} &
\textbf{Max output} & \textbf{Attempts/read timeout} \\
\midrule
Gemini-3-Flash-Preview & \texttt{gemini-3-flash-preview} &
Provider default & 0.2 & Provider default & 3 / 300 s \\
Qwen3-VL-Plus & \texttt{qwen3-vl-plus} &
Binary control unset & 0.2 & Provider default & 3 / 300 s \\
Gemma-4-31B-IT & \texttt{gemma-4-31b-it} &
Unsupported & 0.2 & Provider default & 3 / 300 s \\
Kimi-K2.5 & \texttt{kimi-k2.5} &
Binary control unset & 0.2 & Provider default & 3 / 300 s \\
GPT-5.2 & \texttt{gpt-5.2} &
Provider default & 0.2 & Provider default & 3 / 300 s \\
Claude-Sonnet-5 & \texttt{claude-sonnet-5} &
Provider default & Not sent & 32,768 tokens & 2 / 400 s \\
\bottomrule
\end{tabularx}
\caption{Model identifiers and request controls.  Provider-default output
limits are not replaced by a common artificial cap.  Access timestamps and
the resolved request configuration are retained per request.}
\label{tab:app-model-controls}
\end{table*}

The common transport configuration uses a 10-second connect timeout,
60-second write timeout, 30-second pool timeout, and observable exponential
backoff from 2 to 10 seconds.  Claude-Sonnet-5 receives a longer read timeout
and fewer attempts because a single generation can exceed the default
300-second window; other models retain the common policy.  Transport retries
do not increase the benchmark's LLM-call counter: the counter advances once
per logical model request, while HTTP-attempt telemetry is reported
separately.

Model aliases can be mutable at the provider.  We therefore define the access
time as the recorded request-trace \texttt{started\_at} timestamp rather than a
filesystem modification time.  Each released request trace contains the
model identifier, gateway endpoint, start and completion timestamps,
temperature, maximum-output setting, reasoning-control keys, timeout policy,
HTTP-attempt outcomes, latency, and token usage when returned by the provider.
Credentials are never included.  A release-level manifest maps every table
row to its constituent session IDs and access timestamps, allowing future
changes behind an alias to be distinguished from the evaluated snapshot.

\subsection{Frozen Run Inclusion Rule}
\label{app:run-inclusion}

The repository contains pilot and debugging sessions in addition to paper
runs.  Paper aggregation is therefore defined by configuration fields, not by
placing every file under one directory into the denominator.  A primary-run
episode is eligible for aggregation only when its frozen session manifest
matches all of the following:
\begin{itemize}
    \item \texttt{skill\_mode=no\_skill};
    \item at most three LLM calls and three physics attempts;
    \item a 12-second simulation window sampled at 4 fps;
    \item uniform selection of four frames, packaged as a contact sheet;
    \item the failure-marker cutoff specified in
    Section~\ref{app:temporal-feedback};
    \item the common prompt and deterministic validator versions; and
    \item the declared strict or loose budget map for the condition.
\end{itemize}
Sessions generated with earlier execution windows, alternative feedback
modes, partial level subsets, ad hoc prompts, or skill injection are pilot
runs and are excluded.  Before aggregation, the report generator verifies one
record per model--level--condition key, the expected level count, and the
presence of the associated trajectory.  This rule prevents an incomplete
rerun or a copied session directory from silently changing a paper result.

\subsection{Recorded Artifacts and Metric Reconstruction}
\label{app:metric-reconstruction}

The experiment output is hierarchical.  A session-level
\texttt{summary.json} records the frozen controller configuration and a copy
of all level results.  Each level directory contains \texttt{result.json} and
\texttt{trajectory.json}.  The former stores final outcome, budgets, calls,
physics attempts, material usage, timestamps, and aggregated HTTP telemetry;
the latter stores one record per logical LLM call, including the response,
parsed topology, feedback status, evidence metadata, and per-phase timing.
Round directories retain decoded designs, encoded saves, editor screenshots,
raw physics frames, and repair contact sheets.

All headline metrics can be reconstructed from \texttt{trajectory.json}
without trusting a precomputed table.  The report script restricts events to
the three-call horizon, derives valid and successful call indices, derives
the Recovery@3 eligibility and success flags, then aggregates the resulting
level records.  Model-authored \texttt{thought\_process} strings are not used
to decide validity, success, recovery, cost, or error category.

\begin{table*}[!t]
\centering
\scriptsize
\setlength{\tabcolsep}{3.2pt}
\begin{tabular}{@{}l cc cc cc cc@{}}
\toprule
& \multicolumn{2}{c}{\textbf{Anchor/connect.} ($N=59$)}
& \multicolumn{2}{c}{\textbf{Terrain/geometry} ($N=64$)}
& \multicolumn{2}{c}{\textbf{Dynamic load} ($N=26$)}
& \multicolumn{2}{c}{\textbf{Compositional} ($N=40$)} \\
\cmidrule(lr){2-3}\cmidrule(lr){4-5}\cmidrule(lr){6-7}\cmidrule(lr){8-9}
\textbf{Model} & \textbf{V@3} & \textbf{S@3}
& \textbf{V@3} & \textbf{S@3}
& \textbf{V@3} & \textbf{S@3}
& \textbf{V@3} & \textbf{S@3} \\
\midrule
Gemini & 49.2 & 42.4 & 32.8 & 26.6 & 46.2 & 11.5 & 20.0 & 5.0 \\
Qwen   & 62.7 & 5.1  & 40.6 & 0.0  & 73.1 & 0.0  & 85.0 & 0.0 \\
Gemma  & 52.5 & 35.6 & 34.4 & 6.2  & 61.5 & 11.5 & 62.5 & 0.0 \\
Kimi   & 44.1 & 16.9 & 25.0 & 3.1  & 34.6 & 0.0  & 10.0 & 0.0 \\
GPT    & 54.2 & 13.6 & 43.8 & 7.8  & 34.6 & 3.8  & 20.0 & 0.0 \\
Claude & 94.9 & 71.2 & 85.9 & 50.0 & 96.2 & 61.5 & 77.5 & 15.0 \\
\bottomrule
\end{tabular}
\caption{Strict-condition performance by coarse capability group.  V@3 and
S@3 denote Valid@3 and Success@3; all entries are percentages.  Group sizes
are shown in the headers so that each percentage has an explicit denominator.}
\label{tab:app-strict-groups}
\end{table*}

\begin{table*}[!t]
\centering
\scriptsize
\setlength{\tabcolsep}{4.0pt}
\begin{tabular}{@{}l c c c c c c@{}}
\toprule
& \multicolumn{3}{c}{\textbf{Strict condition}}
& \multicolumn{3}{c}{\textbf{Loose condition}} \\
\cmidrule(lr){2-4}\cmidrule(lr){5-7}
\textbf{Model}
& \textbf{Recovered/eligible} & \textbf{Calls$_{\rm all}$}
& \textbf{Cost$_{\rm succ}$ (\$k; $N$)}
& \textbf{Recovered/eligible} & \textbf{Calls$_{\rm all}$}
& \textbf{Cost$_{\rm succ}$ (\$k; $N$)} \\
\midrule
Gemini & 1/17  & 2.79 & 9.7 (47)  & 10/30 & 1.47 & 16.5 (160) \\
Qwen   & 1/63  & 2.97 & 7.0 (3)   & 16/148 & 2.87 & 15.8 (23) \\
Gemma  & 0/29  & 2.87 & 9.7 (28)  & 25/66 & 2.47 & 18.7 (87) \\
Kimi   & 1/15  & 2.94 & 6.9 (12)  & 7/25 & 2.81 & 13.3 (35) \\
GPT    & 0/17  & 2.95 & 8.0 (14)  & 7/27 & 2.71 & 21.0 (64) \\
Claude & 20/91 & 1.66 & 10.0 (96) & 31/62 & 1.49 & 16.9 (154) \\
\bottomrule
\end{tabular}
\caption{Recovery and resource accounting under strict and loose conditions.
Calls$_{\rm all}$ averages logical LLM calls over all 189 episodes, including
failures.  Cost$_{\rm succ}$ is the mean realized material cost over successful
episodes, in thousands of dollars; the parenthesized value is the number of
successful episodes contributing to that mean.}
\label{tab:app-recovery-resources}
\end{table*}

\section{Additional Quantitative Results}
\label{app:additional-results}

This section reports two compact diagnostics beyond the suite-level results
in the main paper.  To avoid unstable percentages from families containing
only two to six levels, Table~\ref{tab:app-strict-groups} aggregates the 27
families into four mutually exclusive, pre-defined capability groups.  The
strict condition is used because it is the primary benchmark setting.  Model
names are shortened in the tables; their exact API identifiers appear in
Table~\ref{tab:app-model-controls}.

\subsection{Coarse-Grained Capability Breakdown}
\label{app:coarse-capability-results}

The four groups are Anchor and Connectivity (T1B--T1H), Terrain and Geometry
(T2A--T2K), Dynamic Load (T4B--T4D), and Compositional Constraints (G1--G6).
Their sample sizes are 59, 64, 26, and 40 levels, respectively.  These groups
are used only for descriptive diagnosis; they are not additional test splits
and are not used to claim statistical significance.

Claude has the highest Success@3 in every group, while Gemini is the strongest
remaining model on Anchor and Connectivity, Terrain and Geometry, and
Compositional Constraints.  The breakdown also sharpens the validity--success
gap.  For example, Qwen reaches 73.1\% Valid@3 on Dynamic Load and 85.0\% on
Compositional Constraints, but records no physical success in either group.
Likewise, Gemma reaches 62.5\% validity on the Compositional group without a
successful traversal.  Passing deterministic checks therefore remains
insufficient evidence of dynamic functionality across qualitatively different
structural conditions.

\subsection{Recovery and Resource Use}
\label{app:recovery-resource-results}

Table~\ref{tab:app-recovery-resources} reports the Recovery@3 numerator and
denominator, LLM calls averaged over all 189 episodes, and the realized
material cost averaged over successful episodes.  A cost cell has the form
``mean cost in thousands of dollars (number of successful episodes).''  It
reports the material used by the successful design, not the level's initial
budget limit and not a cost-to-limit ratio.

Recovery must be interpreted with its conditional denominator.  Claude
recovers on 20 of 91 eligible strict episodes and 31 of 62 eligible loose
episodes, whereas Qwen's loose rate is based on a much larger set of 148
eligible episodes.  Calls$_{\rm all}$ includes failed episodes that exhaust the
three-call allowance, so it complements the success-conditioned interaction
measure in the main paper.  Cost$_{\rm succ}$ is also conditional: Qwen's low
strict mean, for example, is based on only three successful levels and must not
be read as a standalone efficiency ranking.  Per-level records and trajectories
are included in the artifact inventory in
Appendix~\ref{app:artifact-inventory}.

\setcounter{section}{5}
\section{Paired Three-Round Repair Traces}
\label{app:paired-repair-traces}

This section makes the trajectory-level meaning of Recovery@3 concrete with
two deliberately selected Claude-Sonnet-5 traces.  Both come from the same
strict-condition session and the same G1 rock-clearance family, use a 14~m
span with Road, Wood, and Steel available, and consume three model calls and
three physics attempts.  Every candidate in both traces passes deterministic
validation and enters the native simulator, so neither trajectory contains a
validator-repair round.  The examples are illustrative rather than
representative: they contrast one recovered episode with one episode that
exhausts the interaction budget, and they do not estimate the marginal causal
effect of temporal frames.

\subsection{Paired Case Context}
\label{app:paired-case-context}

The successful trace is
\path{G1_Clearance_Multi2_14m_R2m}, which contains two vehicles with weight
3.0 each.  The unsuccessful trace is
\path{G1_Clearance_HeavyW3_14m_R2m}, which contains one vehicle with weight
9.0.  Table~\ref{tab:app-case-overview} summarizes how the two episodes enter
the Recovery@3 accounting from Section~\ref{app:metric-edge-cases}.  Here, $F$
denotes a valid topology followed by \texttt{failure\_or\_timeout}, and $S$
denotes simulator-confirmed success.

\begin{center}
\centering
\scriptsize
\setlength{\tabcolsep}{3.8pt}
\begin{tabular}{@{}l c c c@{}}
\toprule
\textbf{Case} & \textbf{Load} & \textbf{Outcomes} & \textbf{Recovery@3} \\
\midrule
Multi2
& $2\times3.0$
& $F\!\rightarrow\!F\!\rightarrow\!S$
& Recovered \\
HeavyW3
& $1\times9.0$
& $F\!\rightarrow\!F\!\rightarrow\!F$
& Not recovered \\
\bottomrule
\end{tabular}
\captionof{table}{Overview of the paired three-round traces.  Both episodes are
Recovery@3-eligible after their first physical failure.  Multi2 contributes to
both the numerator and denominator, whereas HeavyW3 contributes only to the
denominator.}
\label{tab:app-case-overview}
\end{center}

\begin{figure*}[t]
\centering
\includegraphics[width=\textwidth]{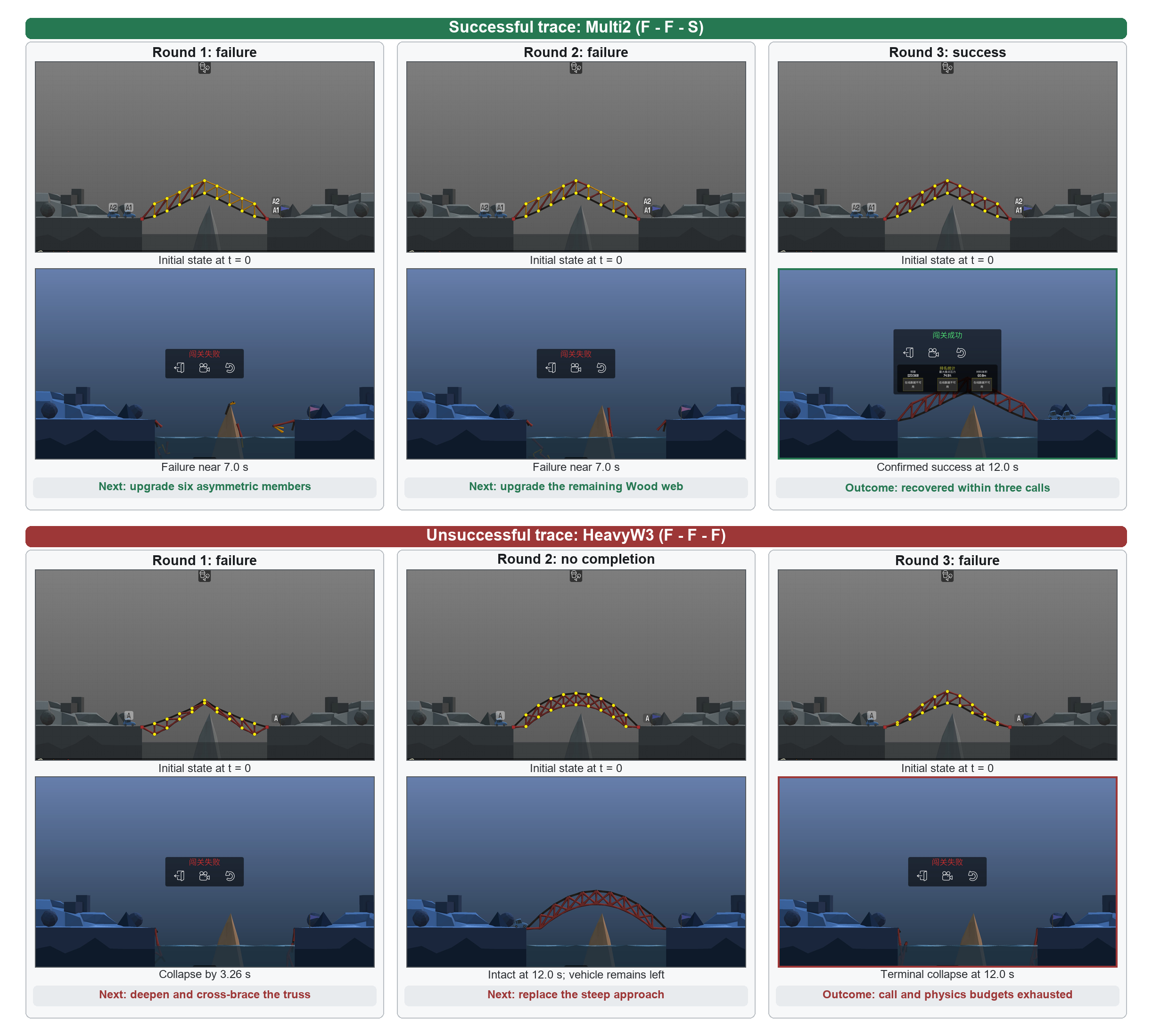}
\caption{Two three-round repair traces under the standard feedback package.
The upper row shows Multi2 progressing from two valid physical failures to
simulator-confirmed success.  The lower row shows HeavyW3 changing failure
signatures across two redesigns but ending in terminal collapse.  Each panel
pairs the candidate at $t=0$ with selected temporal evidence or the terminal
frame.  Action labels are reconstructed from stored design differences and
model-authored decision summaries; they are not causal annotations.}
\label{fig:app-paired-traces}
\end{figure*}

Figure~\ref{fig:app-paired-traces} pairs the initial state of every executed
candidate with a diagnostic or terminal frame from the corresponding rollout.
The edit labels summarize the stored topology difference and the model-written
decision summary.  They describe the action taken by the model, not a verified
ground-truth diagnosis of the physical failure.

\subsection{Successful Trace: Progressive Strengthening}
\label{app:successful-repair-trace}

Multi2 begins with a fixed 20-node, 37-edge geometry containing ten Road
segments and a non-road web of 21 Wood and six Steel members.  The first
rollout remains visually coherent for several seconds before a late failure
near the right side.  The model treats a left--right strength asymmetry as a
hypothesis and changes six apex and right-side Wood members to Steel.  The
second candidate retains every node position and edge endpoint, now with 15
Wood and 12 Steel web members, but produces a similar late physical failure.

For the third call, the model again preserves the complete geometry and
upgrades the remaining fifteen Wood chord and vertical members.  The resulting
candidate has ten Road and 27 Steel members and carries both vehicles to their
targets.  This $F\!\rightarrow\!F\!\rightarrow\!S$ trajectory is therefore a
successful Recovery@3 case: the episode recovers under the complete feedback
package described in Section~\ref{app:temporal-feedback}.  It does not by
itself show that the model's stated explanation was correct or that temporal
frames alone caused the recovery.

\subsection{Failed Trace: Non-Monotonic Redesign}
\label{app:failed-repair-trace}

HeavyW3 initially uses a 20-node, 37-edge design whose ten Road segments are
supported by 27 Steel members.  The first rollout visibly collapses by
approximately 3.26~s.  In response, the model changes the suspended-node
geometry and expands the structure to 45 edges, producing a deeper and more
densely cross-braced Steel truss.  The redesigned bridge remains visibly
intact through the 12~s window, but the vehicle remains near the left approach
and the controller does not confirm task completion.

The model then hypothesizes that the steep entry profile, rather than global
structural instability, prevented completion.  Its third candidate replaces
the road profile with a gentler arch and reduces the topology to 37 edges.
The final execution collapses and terminates in failure, exhausting both the
three-call and three-physics-attempt allowances.  Thus the feedback changes the
observed failure signature and the model's repair direction, but the edits are
not monotonically improving.  Together with the successful trace, this case
illustrates why deterministic validity, physical success, and recovery are
reported separately, while retaining the causal qualifications in
Appendix~\ref{app:claim-scope}.

\FloatBarrier

\setcounter{section}{6}
\section{Extended Limitations and Artifact Documentation}
\label{app:limitations-artifacts}

\subsection{Claim Scope}
\label{app:claim-scope}

PolyBridgeBench evaluates executable structural synthesis in a controlled
two-dimensional game environment.  It is not a civil-engineering design or
certification system.  The benchmark supports the claim that a generated
graph satisfies the stated game constraints and succeeds under the native
simulator; it does not establish real-world safety, code compliance, fatigue
life, constructability, or reliability under unmodeled loads.  The central
scientific object is the gap between deterministic rule compliance and
dynamic functionality, with bridge construction serving as a compact,
falsifiable testbed.

The standard post-failure observation is a package containing the failure
status, current topology, editor image, temporal contact sheet, and interaction
history.  Recovery@3 measures recovery under this complete package.  Unless a
paired feedback ablation is reported, it does not identify the marginal causal
effect of temporal frames, nor does a model-written failure explanation prove
that the model inferred the physical cause correctly.

\subsection{Threats to Validity}
\label{app:threats-validity}

\begin{table*}[t]
\centering
\scriptsize
\setlength{\tabcolsep}{3.5pt}
\begin{tabularx}{\textwidth}{@{}l X X@{}}
\toprule
\textbf{Threat} & \textbf{Protocol control} & \textbf{Residual limitation} \\
\midrule
Simplified physics & All candidates are executed in the same native simulator
with a common success detector. & Two-dimensional game materials omit
continuum behavior, construction sequence, safety factors, and many real
failure mechanisms. \\
Finite procedural families & Parameters vary systematically across 27
families and a separate compositional stratum. & The suite does not establish
open-world or strict held-out-family generalization. \\
Structured environment input & Every model receives the same anchors,
materials, loads, budgets, terrain affordances, and clearance polygons. &
Coordinates reduce the need for pure visual localization; multimodal claims
apply to the complete image-plus-structure package. \\
Human budget calibration & Limits are fixed before model evaluation and only
the scalar budget is exposed. & A manually constructed reference can encode
designer bias and need not be the minimum feasible cost. \\
Approximate rock geometry & One deterministic mapping is used for every model
and checked by a geometric validator. & The triangle approximates a sprite
rather than reading the game's collision mesh. \\
Validator assistance & Invalid outputs consume calls, and the validator never
predicts stability or generates members. & Later performance reflects a
model-plus-feedback protocol, not unaided one-shot model ability. \\
Interface-based success detection & The window, screen coordinate, threshold,
and raw frames are fixed and logged. & Resolution, UI theme, focus, or delayed
completion can cause detector error; non-success is therefore labeled
\texttt{failure\_or\_timeout}. \\
Stochastic and mutable APIs & Request parameters, timestamps, aliases, and
attempt traces are retained. & One trajectory per model--level condition does
not estimate sampling variance, and provider aliases can change after access. \\
Family cue in episode name & Separate family IDs, tags, and generator
parameters are withheld as structured fields. & The model-visible
\texttt{levelName} is human-readable and can contain a T/G family prefix; the
results should not be interpreted as blind family recognition or strict OOD
generalization. \\
Concise decision summaries & The prompt requests only a brief
\texttt{thought\_process}; scoring ignores it. & This field is a model-authored
visible trace, not privileged hidden reasoning and not evidence of a faithful
causal chain. \\
\bottomrule
\end{tabularx}
\caption{Primary threats to validity, controls, and residual limitations.}
\label{tab:app-validity-threats}
\end{table*}

\paragraph{Coverage and generalization.}
The 189 levels cover span, slope, bank asymmetry, bottom anchors, center
platforms, clearance, material restriction, multiple vehicles, and elevated
loads, but they remain a finite sample from hand-designed generator families.
The Compositional Challenge Suite combines known constraint types; it is not a
training/test OOD split.  Family-level results diagnose robustness across
parameterized configurations and should not be described as proof that a
model learned a transferable engineering strategy.

\paragraph{Automation and detector dependence.}
The controller interacts with a desktop game through save-file replacement,
keyboard/mouse automation, screenshots, and fixed-pixel UI detection.  This
choice preserves native physics but introduces dependencies on game version,
screen resolution, interface layout, input focus, and rendering latency.  A
12-second window makes the criterion reproducible but merges slow completion
with physical failure.  Raw frames and detector pixels are therefore retained
so questionable episodes can be audited without relying only on the final
Boolean result.

\paragraph{Budget and clearance calibration.}
Strict budgets are admissibility thresholds derived from successful manual
designs, not certified optima.  A model can construct a cheaper valid bridge,
and no geometric similarity to the reference is required.  Similarly, the
rock triangle is an explicit benchmark convention.  It makes clearance
deterministic but does not claim pixel-perfect equivalence to the simulator's
internal collision shape.  Future versions could replace it with exported
collision geometry and evaluate sensitivity to the polygon boundary.

\paragraph{Provider and sampling variance.}
The primary table reports one trajectory for each model--level--condition key.
It therefore measures the realized behavior of the accessed model snapshot
under a fixed low-temperature protocol, not the expectation over repeated
stochastic samples.  Access timestamps and raw responses make the realization
auditable, but confidence intervals would require repeated complete runs.  In
addition, provider-default reasoning and output limits are not necessarily
matched in internal compute across model families.

\subsection{Artifact Inventory}
\label{app:artifact-inventory}

The intended anonymous artifact separates source, benchmark definitions, and
generated run data as shown in Table~\ref{tab:app-artifact-inventory}.  Paths
are repository-relative.  Local game paths, API credentials, caches, and
unrelated pilot outputs are excluded.

\begin{table*}[t]
\centering
\small
\setlength{\tabcolsep}{4pt}
\begin{tabularx}{\textwidth}{@{}l X X@{}}
\toprule
\textbf{Artifact} & \textbf{Contents} & \textbf{Reproducibility role} \\
\midrule
\texttt{generate/main.py} & Family generators, semantic environment export,
and benchmark serialization. & Regenerates level geometry and metadata. \\
\path{generate/benchmark_levels_v2_money/} & Strict-budget outer,
inner, and context JSON triplets for 189 levels. & Frozen primary benchmark
inputs. \\
\path{generate/benchmark_levels_v2/} & Geometry-matched triplets with
the uniform loose budget. & Diagnostic loose-budget inputs. \\
\texttt{generate/money.json} & Level-to-calibrated-budget mapping. & Rebuilds
strict constraints and successful cost ratios. \\
\texttt{main.py} & Prompt, controller loop, validation, resource accounting,
and run configuration. & Defines logical calls and episode transitions. \\
\texttt{tool/encode.py} and related save tools & Native save encoding,
loading, execution, and feedback-frame processing. & Connects structured
graphs to native simulation. \\
\path{gen/<session>/<model>/<level>/} & Results, trajectories, decoded and
encoded designs, screenshots, raw frames, and contact sheets. & Supports
episode replay and claim auditing. \\
\texttt{result\_stats.py} & Deterministic metric reconstruction and report
generation. & Recreates Valid, Success, Recovery, error, and resource tables. \\
\bottomrule
\end{tabularx}
\caption{Artifact components and their roles.}
\label{tab:app-artifact-inventory}
\end{table*}

\subsection{Release Manifest and Integrity Checks}
\label{app:release-manifest}

The artifact root includes a machine-readable manifest containing the source
commit, benchmark and budget-map hashes, prompt and validator hashes, game
version, operating system, display resolution, Python and package versions,
model aliases, gateway endpoints, access timestamps, and the session IDs used
for every reported model--condition row.  Each session entry records the
expected number of levels and the frozen configuration predicate from
Section~\ref{app:run-inclusion}.  File hashes cover the per-level result and
trajectory files; raw images may be packaged in a separate archive with its
own checksum.

The manifest is considered valid only when it contains no duplicate
model--level--condition keys, missing benchmark levels, missing trajectories,
inconsistent budget maps, or sessions whose call limit, physics limit,
execution window, feedback mode, or skill mode differs from the declared
paper protocol.  Aggregate CSV and LaTeX tables are treated as derived
artifacts: the JSON trajectories, not manually edited tables, are the source
of truth.

Secrets and machine-specific configuration are removed before release.
Specifically, \texttt{.env}, API keys, local Poly Bridge layout paths, and
provider authentication headers are never included.  The benchmark does not
require personal data or human-subject annotations.  Poly Bridge itself is a
proprietary dependency and is not redistributed; reproducing native execution
requires a legally obtained compatible installation.  Generated structured
levels, prompts, controller code, metrics, and non-proprietary audit metadata
can be released independently subject to the final artifact license.

\end{document}